\documentclass[lettersize,journal]{IEEEtran}
\usepackage{graphicx}
\usepackage{booktabs}
\usepackage{amsmath,amsfonts,amsthm,amssymb}
\usepackage{algorithmic}
\usepackage{algorithm}
\usepackage{array}
\usepackage[caption=false,font=normalsize,labelfont=sf,textfont=sf]{subfig}
\usepackage{textcomp}
\usepackage{stfloats}
\usepackage{url}
\usepackage{verbatim}
\usepackage{graphicx}
\usepackage{CJKutf8}
\usepackage{tabularx}
\usepackage{caption}
\usepackage{utfsym}
\usepackage{cite}
\usepackage{lineno} 
\usepackage[backref=false,bookmarks=true]{hyperref}
\usepackage[T1]{fontenc}
\usepackage{soul, color, xcolor} 
\usepackage{ulem} 
\usepackage{color,colortbl}
\usepackage{xcolor}
\usepackage{soul}

\usepackage{pifont}
\newcommand{\cmark}{\ding{51}} 
\newcommand{\xmark}{\ding{55}} 
\usepackage{threeparttable}
\usepackage{multirow}
\usepackage{multicol}
\usepackage{color,colortbl}
\definecolor{Gray}{gray}{0.90}
\newcolumntype{a}{>{\columncolor{Gray}}c}
\definecolor{lightgreen}{RGB}{100,220,100}
\definecolor{darkgreen}{RGB}{30,150,30}
\usepackage{tikz}
\usetikzlibrary{positioning}
\usepackage{changepage}
\usepackage{sidecap}
\usepackage{booktabs}
\usetikzlibrary{calc}
\usetikzlibrary{shapes, fit, backgrounds, decorations.pathreplacing}
\usepackage{ragged2e}

\begin{document}

\title{LFM: Leveraging Foundation Models for Source-Free Universal Domain Adaptation}
\author{
Jing Li, Pan Liu, Meng Zhao, 
Wanli Xue, 
Yanhong Yang, 
Xu Cheng, 
Fan Shi, 
Jianhua Zhang,
\\
Qinghua Hu,~\IEEEmembership{Senior Member,~IEEE,}
and Shengyong Chen,~\IEEEmembership{Senior Member,~IEEE}

\thanks{
Jing Li, Pan Liu, Meng Zhao, Wanli Xue, Yanhong Yang, Xu Cheng, Fan Shi, Jianhua Zhang, and Shengyong Chen are with the School of Computer Science and Engineering, Tianjin University of Technology, Tianjin 300384, China, and also with the Engineering Research Center of Learning-Based Intelligent System, Ministry of Education of the People’s Republic of China, Tianjin University of Technology, Tianjin 300384, China.
(e-mail:  
jing\underline{ }li@tju.edu.cn;
panliu@stud.tjut.edu.cn;
zh\underline{ }m@tju.edu.cn;
xuewanli@email.tjut.edu.cn;
yyh@email.tjut.edu.cn;
xu.cheng@ieee.org;
shifan@email.tjut.edu.cn;
zjh@ieee.org;
sy@ieee.org).

Qinghua Hu is with the School of Artificial Intelligence, Tianjin University, Tianjin 300350, China, also with the Engineering Research Center of City Intelligence and Digital Governance, Ministry of Education of the People’s Republic of China, Tianjin University, Tianjin 300350, China (e-mail: huqinghua@tju.edu.cn).
}
}

\maketitle
\begin{abstract}
Source-free universal domain adaptation (SF-UniDA) adapts a pre-trained source model to an unlabeled target domain under both covariate and label shifts, without access to source data. However, existing SF-UniDA methods rely on inefficient techniques such as threshold tuning and clustering. Foundation models (FMs), known for their generalization and zero-shot capabilities, remain underexplored in SF-UniDA. In this paper, we propose a framework that leverages foundation models (LFM) for SF-UniDA. We use a vision-language model (VLM) to compute similarities between target samples and text labels, including those for unknown classes generated by prompting a large language model. The label shift type is determined by analyzing the coefficient of variation of a similarity-based sample-level score. Unknown samples are identified using a binary Gaussian mixture model fitted to another similarity-based metric. Under a consensus strategy, the pseudo-labels generated by the VLM are refined by the target model initialized with the pre-trained source model, integrating knowledge from both the source domain and foundation models. Finally, these refined pseudo-labels are used to train the target model. Extensive experiments across all possible label shifts and multiple benchmarks demonstrate the effectiveness and superiority of our proposed LFM framework. Our code is available at https://github.com/iamjingli/LFM.
\end{abstract}

\begin{IEEEkeywords}
Universal domain adaptation, source-free, foundation models, pseudo-labels, sample-wise metric.
\end{IEEEkeywords}

\section{Introduction}
\IEEEPARstart{U}{nsupervised} domain adaptation (UDA) transfers knowledge from labeled source domains to unlabeled target domains and has achieved remarkable success \cite{ZhangDecadeSurvey2022}. Traditional UDA methods focus on addressing covariate shifts between domains, assuming a closed-set scenario where the target domain shares the same label set as the source domain. However, this assumption is often unrealistic in real-world applications, where the target domain may have a different set of categories \cite{jingNENO}.

For clarity, let the label sets of the source and target domains be denoted as $\mathcal{Y}^s$ and $\mathcal{Y}^t$, respectively. Three types of label shifts can occur between these domains: partial-set (PDA, $\mathcal{Y}^s \supset \mathcal{Y}^t$) \cite{cao2019learning}, open-set (OSDA, $\mathcal{Y}^t \supset \mathcal{Y}^s$) \cite{liu2022psdc}, and open-partial-set shift (OPDA, $\mathcal{Y}^s \cap \mathcal{Y}^t \neq \emptyset$, $\mathcal{Y}^s \setminus \mathcal{Y}^t \neq \emptyset$, $\mathcal{Y}^t \setminus \mathcal{Y}^s \neq \emptyset$) \cite{you2019universal}. Methods designed to handle one specific label shift may not be applicable to others \cite{saito2020universal}. Since the target domain is unlabeled, we lack prior knowledge about the type of label shift. To tackle this challenge, universal domain adaptation (UniDA) \cite{chen2022geometric} was proposed to handle all potential label shifts. UniDA aims to identify common category samples while rejecting samples from target-private categories, addressing both covariate and label shifts.

Nevertheless, the reliance on a labeled source domain makes UniDA inefficient and potentially non-compliant with stringent data protection policies \cite{voigt2017eu}. To address this, source-free UniDA (SF-UniDA) \cite{qu2023upcycling} uses a pre-trained source model instead of raw source data. 
However, the inaccessibility of source data renders cross-domain data-based UniDA approaches \cite{chen2022geometric,zhu2023UniAM} inapplicable. Moreover, earlier SF-UniDA methods fail to accommodate all possible label shifts \cite{liang2021umad,kundu2020universal}, limiting their practical applicability. Recent SF-UniDA methods rely on either manual thresholding or clustering-based techniques to distinguish common from target-private categories. However, manually tuning thresholds for all potential label shifts across multiple datasets can be cumbersome and sub-optimal \cite{liang2021umad,fu2020learning}. 
{Clustering-based pseudo-labeling methods \cite{qu2023upcycling,wan2024unveiling}, while effective, are computationally expensive and unstable when applied to high-dimensional data \cite{qu2024lead}.} 
Additionally, LEAD \cite{qu2024lead} revisits old-fashioned feature decomposition techniques \cite{li2018domain} widely used in domain adaptation.

Recently, foundation models (FMs), including large language models (LLMs) and vision–language models (VLMs), have demonstrated impressive zero-shot generalization capabilities by leveraging large-scale semantic knowledge encoded in billions of parameters \cite{bommasani2021opportunities}. Recent studies have shown such capabilities can benefit the adaptation of pre-trained models by providing transferable representations or semantic priors, even under domain gap \cite{wen2024cross,zhang2025source}. While a small number of recent works have begun to incorporate FM-based representations (e.g., CLIP \cite{radford2021learning}) into SF-UniDA-related scenarios, existing approaches \cite{zhang2025source,tang2024source} primarily focus on the close-set setting, and do not explicitly address label-shift type determination and unknown-class separation as first-class challenges under source-free constraints. In particular, how to systematically exploit FMs to support label-shift-aware adaptation and target-specific refinement, rather than using them as direct classifiers, remains underexplored in the SF-UniDA setting.

In this paper, we propose a framework that leverages foundation models (LFM) for SF-UniDA. Specifically, we first utilize the expert knowledge of an LLM to generate text labels for potential target-private categories that are unknown to the source domain. Next, we employ a VLM to perform zero-shot inference using both target images and text labels for both known and unknown categories. Based on the similarities between these modalities, we introduce a sample-level metric score. By analyzing the coefficient of variation of these scores across all target samples, we determine the type of label shift. If the label shift is OSDA or OPDA, we apply a binary Gaussian mixture model (GMM) to identify target-private categories as unknown, and classify others according to the similarities. Otherwise, the pseudo-labels of target samples are directly obtained from CLIP. However, FMs trained for general purposes may not perform optimally on all datasets and tasks. Therefore, we design a module that enables the target model, initialized from the pre-trained source model, to refine the pseudo-labels generated by the VLM through a consensus strategy. These refined pseudo-labels integrate knowledge from both the foundation models and the pre-trained source model. Finally, the target model is adapted using these refined pseudo-labels. We evaluate our framework across all potential label shifts using four popular benchmarks. Extensive experimental results demonstrate the effectiveness and superiority of our method.

The contributions of our work are summarized as follows:
\begin{itemize}
\item[$\bullet$] We propose {an effective and practical framework} that leverages foundation models during training for SF-UniDA, utilizing large language models to generate text labels for unknown classes and vision-language models for zero-shot recognition, enabling the adaptation of a pre-trained source model to the target domain under both covariate and label shifts, without accessing source data. 
\item[$\bullet$] We {develop a unified pipeline} featuring two key modules to tackle label shift detection and knowledge fusion in SF-UniDA. A sample-wise metric and its coefficient of variation are used to infer the label shift type, with a similarity-based GMM applied to identify unknown samples when necessary. Additionally, a pseudo-label refinement module combines CLIP predictions with those of the target model—initialized from the source model—via a confidence-consensus strategy, improving pseudo-label quality and enhancing adaptation performance. 
\item[$\bullet$] Extensive experimental results and analysis, conducted on widely used benchmarks including Office-31, Office-Home, VisDA, and DomainNet, verify the effectiveness and superiority of our proposed approach, {highlighting the potential of foundation-model-assisted adaptation for future research on SF-UniDA}.
\end{itemize}

The remainder of this paper is organized as follows. Section II introduces related works. Section III describes the problem setting and elaborates on our proposed framework. Experimental results and analysis on various benchmarks are in Section IV. Finally, we conclude our work in Section V.

\section{Related Work}  
\subsection{Unsupervised Domain Adaptation}
Unsupervised domain adaptation (UDA) tackles the challenge of distribution shifts between a labeled source domain and an unlabeled target domain. These shifts are commonly referred to as domain or covariate shifts, can significantly degrade model performance on the target data. Existing UDA methods are generally divided into two categories: metric-based approaches, which aim to minimize the distributional discrepancy between the two domains using specific metrics \cite{long2015learning,long2017deep}, and adversarial-based approaches, which employ adversarial training between a feature extractor and a domain discriminator to learn domain-invariant representations \cite{ganin2016domain,DomainPromptTuning}. Although these methods have laid the groundwork for UDA, they typically rely on strict and often unrealistic assumptions. For example, they often assume that the source and target domains share the same category set and can be accessed simultaneously \cite{wang2022cross}, which limits their applicability in more complex real-world scenarios.
\subsection{Source-Free Universal Domain Adaptation}
In recent years, more realistic domain adaptation settings have emerged by relaxing the strict assumptions of UDA. The target domain may have a different label set than the source, leading to various label shift scenarios, such as OSDA~\cite{liu2022psdc,jing2023ICME}, PDA~\cite{cao2019learning}, and OPDA~\cite{you2019universal}. However, methods designed for one specific setting often fail to generalize to others. To address this limitation, universal domain adaptation (UniDA)~\cite{saito2020universal} was proposed to handle all types of label shifts.

Another research direction, source-free domain adaptation (SFDA)~\cite{liang2020we,Tian2024,DPL2025TMM}, focuses on adapting a pre-trained source model to an unlabeled target domain without accessing the source data, which is especially useful in scenarios with storage constraints or privacy concerns. Recent advances in SFDA have explored diverse perspectives, including leveraging inter-sample relations, improving pseudo-label reliability, handling class-imbalance problems, and extending to incremental settings.
For example, HRD~\cite{xing2024hierarchical} exploits the inherent relations among target samples to transfer knowledge from a source-like subdomain to a target-specific subdomain, facilitating robust model training under the source-free setting.
To improve target-domain prediction reliability, DCPL~\cite{eccv2024DCPL} estimates the true class posterior by modeling pseudo-label corruption through a learned noise transition matrix. Beyond the source-free constraint, RNNCDS~\cite{Antonio2024CDS} further considers class distribution variations by constructing class-imbalanced domain adaptation benchmarks and adopting class-wise accuracy as its evaluation metric.
GROTO~\cite{deng2025multi} addresses class-incremental SFDA, where the labeled source domain contains all classes, while unlabeled target data arrive incrementally without access to the source. While our method focuses on object recognition, SF-UT~\cite{Hao2024ECCV} proposes a lightweight source-free adaptation strategy for object detection, and is evaluated using different benchmarks and metrics from ours.

However, most existing SFDA methods do not explicitly address the coexistence of the source-free constraint and target-private classes, which are commonly encountered in practical applications and can substantially degrade adaptation performance~\cite{fang2024source}.
Only a few studies have explored source-free universal domain adaptation (SF-UniDA)~\cite{qu2024lead}, where both label shifts and the absence of source data must be handled simultaneously. Existing approaches either overlook certain shift types~\cite{kundu2020universal} or rely on heuristic thresholding~\cite{liang2021umad} or computationally expensive clustering strategies~\cite{qu2023upcycling}.
In contrast, our method advances SF-UniDA by systematically addressing label shifts and unknown-class challenges under the source-free constraint, while leveraging foundation models for semantic priors rather than task-specific classifiers.

\subsection{Foundation Models}
Foundation models (FMs) are large-scale machine learning models with a massive number of parameters, widely recognized for their outstanding performance across diverse tasks. Large language models (LLMs) like GPT-4 \cite{achiam2023gpt} are trained on massive datasets and show strong reasoning abilities. Multimodal vision-language models (VLMs) such as CLIP \cite{radford2021learning} learn modality-invariant features, enabling remarkable generalization and zero-shot inference. Recent studies have explored the use of FMs,
particularly CLIP, for cross-domain tasks. Deng et al. \cite{deng2023universal} freeze CLIP encoders and distill knowledge via self-calibration to avoid fine-tuning performance drops. CROW \cite{wen2024cross} applies a cluster-then-match strategy in CLIP’s representation space for open-world discovery. Co-learn++ \cite{zhang2025source} improves pseudo-label quality by jointly training the source model and CLIP. Yu et al. \cite{yu2025open} utilize the robustness of CLIP to distribution shifts through entropy-based optimization for OSDA. PromptDIV \cite{PromptDIV2024ICIP} decouples domain-invariant and variant features with domain-specific prompts, enhancing feature alignment in open-set domain adaptation. COSMo \cite{Monga_2024_BMVC} handles open-set multi-target adaptation by separating prompts for known and unknown classes, using source-guided prompt learning with frozen CLIP encoders and a domain-specific bias network to address domain and class shifts. AEM \cite{xiao2024adversarial} guides additional classifiers using CLIP and the target model, employing adversarial learning for knowledge transfer. As a SFDA framework, DIFO \cite{tang2024source} uses frozen CLIP via prompt tuning and mutual information maximization, alternating prompt optimization with regularized knowledge distillation for adaptation.

EOE \cite{pmlr-v235-cao24d} suggests the performance of CLIP in out-of-distribution detection can be enhanced by outlier exposure, which leverages the reasoning abilities of LLMs. Similarly, we incorporate LLM-generated names of potential unknown categories into the text classifier to enhance the accuracy of pseudo-labels assigned by CLIP to target samples. Unlike methods that directly adopt foundation models as inference backbones, our approach exploits their knowledge indirectly through pseudo-labels generated during training. Consequently, the deployed target model remains compatible with ImageNet-pre-trained ResNet-based \cite{he2016deep} DA baselines, enabling inference-time architectural fairness in experimental comparisons. Moreover, while most FM-based methods focus on specific challenges, ours simultaneously addresses source data absence and diverse forms of label shift. 

\begin{figure*}[!th]
\centering
\includegraphics[scale=0.55]{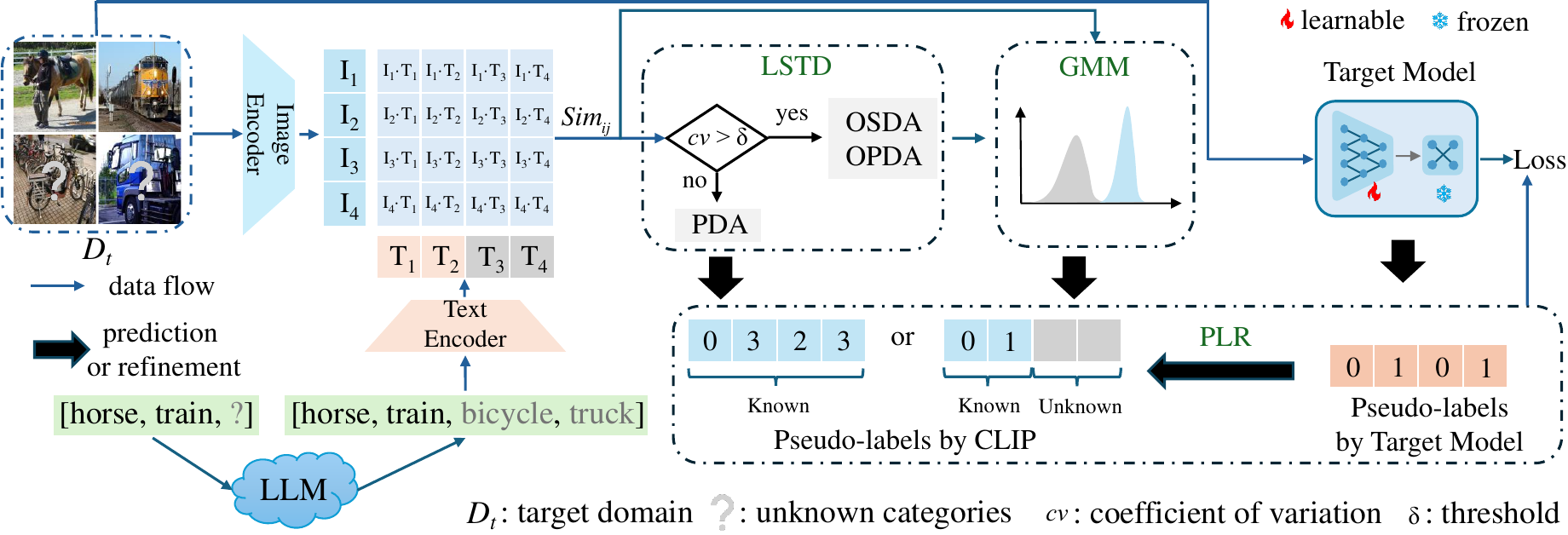}
\caption{Method overview. Known category labels prompt the LLM to generate labels for potential unknowns, allowing CLIP to distinguish them. The LSTD (label shift type determination) module determines the presence of unknown categories, and GMM (Gaussian mixture model) separates them if detected. The PLR (pseudo-label refinement) module refines CLIP pseudo-labels using target model predictions, enabling joint knowledge integration for training.}
\label{fig2}
\vspace{-0.15in}
\end{figure*}

\section{Method}
\subsection{Overview}
Under the SF-UniDA setting, there exists a labeled source domain $D_s=\{(x_i^s, y_i^s)\}_{i=1}^{N_s}$, where $x_i^s \in \mathcal{X}^s, y_i^s \in \mathcal{Y}^s$, and $|\mathcal{Y}^s|= C_s$, as well as an unlabeled target domain $D_t=\{(x_i^t, ?)\}_{i=1}^{N_t}$, where $x_i^t \in \mathcal{X}^t$. For clarity, we refer to the target label set as $\mathcal{Y}^t$, and define $\mathcal{Y}=\mathcal{Y}^s \cap \mathcal{Y}^t$ as the common label set, $\bar{\mathcal{Y}}^s=\mathcal{Y}^s \setminus \mathcal{Y}$ as the source-private label set, and $\bar{\mathcal{Y}}^t=\mathcal{Y}^t \setminus \mathcal{Y}$ as the target-private label set, which is unknown to the source. As mentioned earlier, in addition to domain shift ($\mathcal{X}^s \neq \mathcal{X}^t$), there are three possible types of category shifts: PDA, OSDA, and OPDA. Since the target domain is unlabeled, we have no prior knowledge about the type of label shift we are facing. The goal is to classify target samples that belong to $\mathcal{Y}$, or reject those that belong to $\bar{\mathcal{Y}}^t$. Note that the source domain is unavailable during adaptation and is only used to pre-train the source model. Following previous works \cite{qu2024lead}, the target model, consisting of a feature extractor $f$ and a classifier $g$, is initialized using the source-pretrained model. The classifier $g$ remains frozen, while the feature extractor $f$ is fine-tuned during adaptation.

The overview of our proposed method is illustrated in Fig.~\ref{fig2}. Given a set of text labels for known classes in $\mathcal{Y}^s$ (e.g., `horse, train' in Fig.~\ref{fig2}), we first adopt the designed prompts in EOE \cite{pmlr-v235-cao24d} to envision a set of text labels for potential unknown classes in $\bar{\mathcal{Y}}^t$ (e.g., `bicycle', `truck' in Fig.~\ref{fig2}) using an LLM. Both the known class labels and the generated unknown class labels are then input into the text encoder of CLIP \cite{radford2021learning} to build the textual classifier. For the target samples, we employ CLIP to compute the similarities between their visual features and the textual classifier. Based on these similarities, we propose a sample-specific score that indicates the likelihood of each sample belonging to the common categories. The LSTD module illustrated in Fig.~\ref{fig2} is designed to determine the type of label shift we are facing. If the detected shift is OSDA or OPDA, we separate the known samples from $\mathcal{Y}$ and the unknown samples in $\bar{\mathcal{Y}}^t$ using a binary GMM fitted to the distribution of a similarity-based metric. The unknown samples are labeled as  `unknown', while the known samples are assigned pseudo-labels based on these similarities. If the label shift is PDA, where no unknown categories exist in the target, samples are directly assigned pseudo-labels. To leverage the task-specific knowledge in the target model, we use the pseudo-labels predicted by the model to refine the pseudo-labels initially generated from the similarity calculations by CLIP. This refinement process is referred to as PLR in Fig.~\ref{fig2}. Finally, the refined pseudo-labels are used to train the target model. The remainder of this section elaborates on each module of the proposed method.

It is important to emphasize that, although FMs are leveraged in our framework, they are not used as classifiers for target prediction. Instead, they serve as auxiliary semantic priors that facilitate pseudo-label generation and refinement under distribution mismatch.
Specifically, the LLM is only used offline to generate candidate textual labels for potential unknown classes conditioned on the known-class set, while frozen CLIP provides cross-modal similarities utilized in the label shift type determination. All subsequent adaptation and inference are performed by the ResNet-based target model adapted from the source model, without requiring any FMs. This design differs from methods that directly rely on FMs for classification. Our framework explicitly models label shifts, unknown-class separation, and target-domain uncertainty in a structured manner, enabling robust adaptation beyond the capabilities of direct FM-based prediction.

\subsection{Generating Envisioned Unknown Class Labels} 
We denote the number of possible unknown categories in the target domain as $C_{unk}=\alpha C_s$, where $\alpha$ is a hyperparameter. Given a set of text labels for the known classes in $\mathcal{Y}^s$, we prompt a LLM to envision a corresponding set of text labels for the unknown classes in $\mathcal{Y}^t$. The objective is to obtain semantically plausible yet non-overlapping textual labels for potential unknown categories, which are subsequently used for vision--language alignment in CLIP. We adopt prompt templates from EOE \cite{pmlr-v235-cao24d}, which are designed to anchor the generation of unknown class labels on visual similarity. Concretely, the LLM is instructed to first summarize broad visual concepts shared by the known classes, and then to envision a set of categories that visually resemble these concepts but are not directly related to them. To ensure brief and readable outputs, we first constrain the LLM response format as follows.
\begin{figure}[!h]
    \centering
    \begin{tikzpicture}
        \small
        \definecolor{chatcolor1}{HTML}{5fedb7} 
        \definecolor{shadecolor}{gray}{0.9}
        \definecolor{promptcolor}{HTML}{D6EAF8}  
        \fontfamily{cmss}\selectfont
        
        \node[align=left, text width=0.45\textwidth, fill=shadecolor, rounded corners=1mm, anchor=north west] (node1) at (0,0) 
        {
        Before my question, I will give you an example; please strictly follow my answer format (use `-' before every item) and only provide the answers without any additional text.
        };       
        \node[draw, black, line width=1pt, rounded corners=1mm, inner sep=4pt, fit=(node1) 
        ] {};

    \end{tikzpicture}
    \caption{LLM prompt to constrain the LLM response format.}
    \label{fig: format prompts}
\end{figure}

\begin{figure}[!h]
    \centering
    \begin{tikzpicture}
        \small
        \definecolor{chatcolor1}{HTML}{5fedb7} 
        \definecolor{shadecolor}{gray}{0.9}
        \definecolor{promptcolor}{HTML}{D6EAF8}  
        \fontfamily{cmss}\selectfont
        
        \node[align=left, text width=0.45\textwidth, fill=shadecolor, rounded corners=1mm, anchor=north west] (node1) at (0,0) {\textbf{Q:} I have gathered images of 4 distinct categories: [`Husky dog', `Garfield cat', `churches', `truck']. Summarize what broad categories these categories might fall into based on visual features. Now, I am looking to identify 5 categories that visually resemble these broad categories but have no direct relation to them. Please list these 5 classes for me.
        \textbf{A:} These 5 classes are: -black stone -mountain -ginkgo tree -river -rapeseed.
        \vspace{1\baselineskip}

        \textbf{Q:} I have gathered images of $C_s$ distinct categories: [\textit{class-info}]. Summarize what broad categories these categories might fall into based on visual features. Now, I am looking to identify $C_{unk}$ categories that visually resemble these broad categories but have no direct relation to them. Please list these $C_{unk}$ classes for me.};
        \node[align=left, text width=0.28\textwidth, fill=chatcolor1, rounded corners=1mm, anchor=north east] (node2) at ([yshift=-0.2cm]node1.south -| node1.east) {\textbf{A:} These $C_{unk}$ classes are:};       
        \node[draw, black, line width=1pt, rounded corners=1mm, inner sep=4pt, fit=(node1) (node2)] {};
    \end{tikzpicture}
    \caption{LLM prompt for unknown-category text generation.}
    \label{fig:unk-prompts}
\end{figure}

As shown in Fig.~\ref{fig:unk-prompts}, we then provide the LLM with an example question–answer pair followed by the actual query.
\textit{class-info} in the prompts denotes the text labels of all known classes in $\mathcal{Y}^s$. Following the enforced output format, the LLM returns a list of $C_{unk}$ candidate text labels, each corresponding to an envisioned unknown category. We then perform an overlap check between the generated unknown labels and the known class labels in $\mathcal{Y}^s$. If any overlap exists, the generated results are discarded and the prompting process is repeated. All generated labels passing the check are directly used as the text labels for the unknown categories, without additional manual filtering. To account for the variability of LLM outputs, we fix the prompt templates and run each task three times using the same prompts. The final experimental results are reported as the average over these runs. Notably, we do not employ multiple prompt variants or prompt ensembling.

\subsection{Computation of the Sample-specific Score}  
Let the total number of categories in the target domain be $C_n=C_s+C_{unk}$, where $C_s$ and $C_{unk}$ denote the numbers of known source classes and unknown target classes, respectively. 
The similarity computed by CLIP is defined as follows:
\begin{equation}
        \begin{aligned}
                Sim_{ij} &= I_i \cdot T_j,
        \end{aligned}
        \label{eq1}
\end{equation}
where $I_i$ denotes the image encoding of the $i$-th target sample $x_i^t$, and $T_j$ denotes the text encoding of the $j$-th possible category in the target domain. The difference between $Sim_{ij}$ and $Sim_{ik}$, where $j \neq k$, is small, making the resulting similarities insufficiently discriminative. To enhance their discriminability, we rescale the similarities as follows:
\begin{equation}
        \begin{aligned}
                aSim_{ij} = Sim_{ij} / t,
        \end{aligned}
        \label{aSim}
\end{equation}  
where $t$ is a scaling factor. Then, we define a sample-specific metric to differentiate between samples of known and unknown classes. The score for sample $x^t_i$ is defined as follows:
\begin{equation}
        \begin{aligned}
                score_i = \mathop{\text{max}}\limits_{j\in [1, C_s]}
                {\frac{e^{aSim_{ij}}}{\sum_{m=1}^{C_n}e^{aSim_{im}}}}\\ - \beta \cdot \mathop{\text{max}}\limits_{k\in (C_s, C_n]}
                {\frac{e^{aSim_{ik}}}{\sum_{m=1}^{C_n}e^{aSim_{im}}}},
        \end{aligned}
        \label{eq3}
\end{equation} 
where the hyperparameter $\beta$ balances the influence of the known and unknown category similarities.

\subsection{Label Shift Type Determination (LSTD)}
If the label shift is PDA, every target sample belongs to a known class. However, if the label shift is OSDA or OPDA, only samples belonging to known categories should be classified into a known category, while unknown samples will be rejected. Therefore, it is essential to first determine the label shift type. Because samples of known classes have a high score while samples of unknown classes have a low score, the distribution of the scores exhibits high variance if the target domain contains unknown categories in the case of OSDA or OPDA. Therefore, we determine the label shift type according to the coefficient of variation ($cv$) \cite{coefficientofvariation} of $score_i$ across all target samples, which is defined as follows:
\begin{equation}
        \begin{aligned}
        cv=\frac{\sigma(\{score_i\}_{i=1}^{N_t})}{E(\{score_i\}_{i=1}^{N_t})},
    \end{aligned}
\end{equation}
where $\sigma(\cdot)$ represents the calculation of the standard deviation, and $E(\cdot)$ represents the calculation of the mean value. 
To enhance the discriminability of $cv$, we only use the $score_i$ whose value ranks top 30\% or bottom 30\% in the target domain. Since $cv$ normalizes variability relative to the mean, it enables comparisons across datasets with different score distributions, thereby allowing a single threshold to be applied across all benchmarks. As shown in Fig.~\ref{fig2}, if $cv$ exceeds the threshold $\delta$, we classify the label shift as OSDA or OPDA; otherwise, it is PDA.

\subsection{Unknown Categories Separation}
\label{section:GMM}
If the label shift is OSDA or OPDA, we must identify samples from target-private categories. Since $score_i$ reflects the likelihood that $x^t_i$ belongs to known classes, we fit a binary Gaussian mixture model (GMM) to these scores, avoiding manual thresholding. As shown in Fig.~\ref{fig2}, samples with lower scores are labeled as ``unknown.''

However, as shown in Fig.~\ref{fig:rateVSscore}, $score_i$ causes the GMM to misclassify some known samples as unknown. We attribute this behavior to the scaling in $aSim_{ij}$ and $score_i$, which intentionally amplifies score differences to aid label shift detection via $cv$. While this polarization facilitates label shift type determination, it negatively affects the GMM’s decision boundary, leading to misclassification of some known samples.

To address this, we define a new sample-specific metric $rate_i$ to replace $score_i$ in the GMM-based unknown class separation:  
\begin{equation}
        \begin{aligned} 
                rate_i &= \mathop{\text{max}}\limits_{j\in [1, C_s]}Sim_{ij} - \beta \cdot \mathop{\text{max}}\limits_{k\in (C_s, C_n]}Sim_{ik},
        \end{aligned}
        \label{rate}
\end{equation}
where $Sim_{ij}$ is given in Eq.~(\ref{eq1}). Like $score_i$, a higher $rate_i$ indicates that $x_i^t$ is more likely to belong to the known classes. We denote the known sample set as $\mathcal{S}_k$ and the unknown set as $\mathcal{S}_u$.

\subsection{Pseudo-Label Refinement (PLR)} \label{PLR}
CLIP demonstrates exceptional zero-shot prediction performance across a variety of datasets \cite{radford2021learning}. Therefore, instead of using a source-pretrained model, we employ CLIP to generate the initial pseudo-labels for target samples, leveraging the vast knowledge embedded in CLIP. However, it is important to note that CLIP may exhibit sub-optimal performance in certain domains or tasks \cite{xiao2024adversarial}. The target model inherits the source knowledge because it is initialized with the source-pretrained model. The final pseudo-labels should integrate knowledge from both CLIP and the target model. Specifically, we propose a strategy to refine the pseudo-labels generated by CLIP with the pseudo-labels generated by the target model. 

Note that samples identified as unknown by GMM do not participate in this refinement. Given a target sample $x_i^{t}$ from the common categories, CLIP generates a pseudo-label $\tilde{y}_i^{clip}$ for it according to $aSim_{ij}$ defined in Eq.~(\ref{aSim}) as follows: 
\begin{equation}
        \begin{aligned}
                \tilde{y}_i^{clip} = \mathop{\text{argmax}}\limits_{j\in [1, C_s]}
                {\frac{e^{aSim_{ij}}}{\sum_{k=1}^{C_s}e^{aSim_{ik}}}}.
        \end{aligned}
        \label{CLIPpseudo-lable}
\end{equation}
The confidence $conf^{clip}_i$ corresponding to $\tilde{y}_i^{clip}$ is defined as follows:
\begin{equation}
        \begin{aligned}
                {conf}^{clip}_i = \mathop{\text{max}}\limits_{j\in [1, C_s]}
                {\frac{e^{aSim_{ij}}}{\sum_{k=1}^{C_s}e^{aSim_{ik}}}}.
        \end{aligned}
        \label{eq5}
\end{equation}
For $x^t_i$, we retain the top-3 confident pseudo-labels generated by CLIP and collect them into a set denoted as ${P_i}$. As to the 2nd and 3rd pseudo-labels, we define them as the labels corresponding to the 2nd and 3rd highest confidence values relative to the highest confidence that is $conf^{clip}_i$.   

As the target model is initialized with the source-pretrained model, it will output a $C_s$-dimensional probability vector $p_i$ for each sample $x^t_i$, which is not accurate due to domain shifts. To improve the reliability of the prediction, we update $p_i$ through a neighborhood soft-voting mechanism \cite{litrico2023guiding} as follows:
\begin{equation}
\bar{p}_i=\frac{1}{K} \sum_{{x^t_j} \in \mathcal{N}_i} p_j,
\label{eq:neigborVote}
\end{equation}
where $\mathcal{N}_i$ denotes a set containing the $K$ nearest neighbors of $x^t_i$ in feature space, based on the cosine distance.

\begin{figure}[!t]
\centering
\includegraphics[scale=0.43]{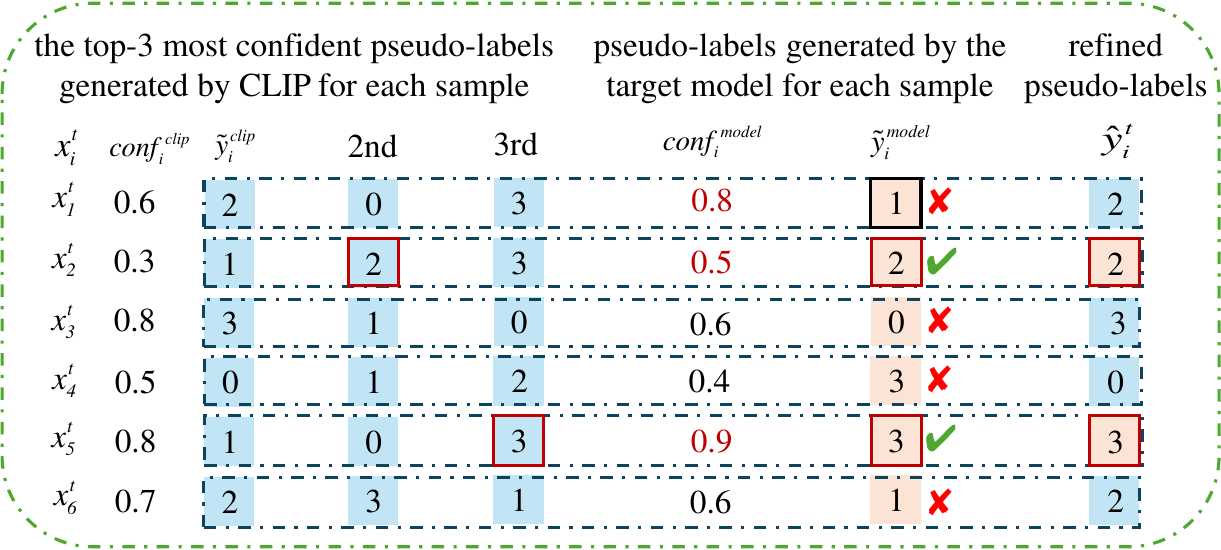}
\caption{Pseudo-Label Refinement. 
The left part shows the top-3 pseudo-label candidates generated by CLIP for each $x^t_i$  (forming the candidate set $P_i$), together with the confidence ${conf}^{clip}_i$ for the prediction $\tilde{y}^{clip}_i$. The middle part presents the pseudo-labels $\tilde{y}^{model}_i$ predicted by the target model for $x^t_i$, along with corresponding confidence ${conf}^{model}_i$. The right part illustrates the refined pseudo-labels $\hat{y}^t_i$, which are determined by jointly considering prediction confidence and cross-model consensus.
}
\label{fig4}
\vspace{-0.15in}
\end{figure}

Similarly, we define the pseudo-label of $x_i^t$ generated by the target model and its corresponding confidence ${conf}_i^{model}$ as follows:
\begin{equation}
        \begin{aligned}
                \tilde{y}_i^{model} = \mathop{\text{argmax}}\limits_{j\in [1, C_s]}\bar{p}_{ij}.
        \end{aligned}
        \label{eq7}
\end{equation}
\begin{equation}
        \begin{aligned}
                {conf}_i^{model} = \mathop{\text{max}}\limits_{j\in [1, C_s]}\bar{p}_{ij}.
        \end{aligned}
        \label{conf_model}
\end{equation}
As shown in Fig.~\ref{fig4}, we utilize the pseudo-labels generated by the target model to refine the pseudo-labels generated by CLIP using a confidence-and-consensus strategy. The strategy is defined as follows:
\begin{equation}
        \begin{aligned}
                \hat{y}^t_i = \begin{cases} \tilde{y}_i^{model},\quad ({conf}^{model}_i > {conf}^{clip}_i) \wedge (\tilde{y}_i^{model} \in {P_i}),\\
                        \tilde{y}_i^{clip},\quad\quad otherwise.
                \end{cases}
        \end{aligned}
        \label{eq8}
\end{equation}
where $\hat{y}^t_i$ denotes the final pseudo-label for $x^t_i$ after the refinement. As illustrated in Fig.~\ref{fig4}, consider a target sample $x^t_1$. The target model predicts label ``1'' with a confidence of 0.8, which is higher than the confidence assigned by CLIP to its predicted label ``2''. However, the model-predicted label ``1'' does not belong to the set ${P_1}=\{2,0,3\}$, which consists of the top-3 most confident pseudo-labels generated by CLIP for $x^t_1$. This indicates that the target model prediction does not agree with CLIP’s high-confidence label candidates. According to Eq.~(\ref{eq8}), although ${conf}^{model}_1 > {conf}^{clip}_1$, the condition $\tilde{y}^{model}_1 \in P_1$ is not satisfied. Therefore, the CLIP pseudo-label ``2'' is retained, i.e., $\hat{y}^t_1 = \tilde{y}^{clip}_1$, rather than being replaced by the target model prediction.  

In short, $\tilde{y}_i^{clip}$ will only be replaced by $\tilde{y}_i^{model}$ if ${conf}^{model}_i$ is greater than ${conf}^{clip}_i$ (the confidence matters) and $\tilde{y}_i^{model}$ is in set ${P_i}$ (the consensus between the models matters).

\subsection{Training and Inference}
For the samples in the known categories set $\mathcal{S}_k$, we use pseudo-labels to supervise the training of the target model via cross-entropy loss, as defined in Eq.~(\ref{loss:known}). For the samples in the unknown categories set $\mathcal{S}_u$, we apply a self-entropy maximization loss, defined in Eq.~(\ref{loss:unknown}), to train the model, encouraging uncertainty predictions for the target-private category samples.

\begin{equation}
        \begin{aligned}
                \mathcal{L}_{k} = -\frac{1}{|\mathcal{S}_k|}\sum_{i=1}^{|\mathcal{S}_k|}\sum_{j=1}^{C_s}\hat{y}_i^t \log p_{ij}.
        \end{aligned}
        \label{loss:known}
\end{equation}
\begin{equation}
        \begin{aligned}
                  \mathcal{L}_{unk} =\frac{1}{|\mathcal{S}_u|}\sum_{i=1}^{|\mathcal{S}_u|}\sum_{j=1}^{C_s}p_{ij} \log p_{ij},
        \end{aligned}
        \label{loss:unknown}
\end{equation}
where $p_{ij}$ represents the probability of the $j$-th class for the target sample $x^t_i$ predicted by the target model. The overall training loss is defined as: $\mathcal{L} = \mathcal{L}_{k} + \mathcal{L}_{unk}$.

During inference, only the target model is used. Following the strategy proposed by GLC \cite{qu2023upcycling}, we use the normalized Shannon entropy to measure the uncertainty of the target model on each sample: 
\begin{equation}
        H(x^t_i) = - \frac{1}{logC_s}\sum_{j=1}^{C_s}p_{ij} \log p_{ij}.
        \label{eq12}
\end{equation}
The predicted category for $x^t_i$ is then determined as follows: 
\begin{equation}
                y(x^t_i) = \begin{cases} unknown, & H(x^t_i) \geq w,\\
                        \mathop{\textit{argmax}}\limits_{j\in [1, C_s]} p_{ij}, & H(x^t_i) < w,
                \end{cases}
        \label{eq13}
\end{equation}
where the threshold $w$ is a hyperparameter.

\section{Experiments}
\subsection{Datasets}
Office-31 \cite{saenko2010adapting} contains 31 categories across 3 domains: Amazon (A), DSLR (D), and Webcam (W). Office-Home \cite{venkateswara2017deep} is a challenging benchmark with 65 categories and 4 domains: Artistic (Ar), ClipArt (Cl), Product (Pr), and Real-world (Rw). VisDA \cite{peng2017visda} includes 12 categories, with synthetic images as the source domain and real images as the target one. DomainNet \cite{peng2019moment} is a large-scale benchmark featuring 345 categories, and Painting (P), Real (R), and Sketch (S) are used.

\begin{table}[t]\renewcommand\arraystretch{1.2}
        \centering
        \scalebox{1.1}{
                \begin{tabular}{l|ccc}
                        \toprule
                        \multirow{2}{*}{Dataset} & \multicolumn{3}{c}{Class Split($\mathcal{Y}/ \bar{\mathcal{Y}}^s/ \bar{\mathcal{Y}}^t$)}  \\
                        \cmidrule{2-4} & OPDA  & OSDA  & PDA \\ 
                        \midrule
                        Office-31 & 10/10/11 & 10/0/11 & 10/21/0 \\
                        Office-Home & 10/5/50 & 25/0/40 & 25/40/0 \\
                        VisDA & 6/3/3 & 6/0/6  & 6/6/0  \\
                        DomainNet & 150/50/145   & -       & - \\ 
                        \bottomrule
                \end{tabular}
        }
        \caption{Details of class split. 
        ``-'' denotes ``not applicable''.
        }
        \label{tab:label_split}
        \vspace{-0.1in}
\end{table}

\subsection{Implementation Details}
We follow the standard pre-training protocol of prior source-free methods \cite{liang2020we}. The labeled source domain trains a source model with feature extractor $f_s$ and classifier $g_s$. We use ResNet-50 for Office-31, Office-Home, DomainNet, and ResNet-101 for VisDA, initialized with ImageNet-1k weights and followed by a 2-layer weight-normalized linear classifier. For adaptation, we use SGD (momentum 0.9), a fixed learning rate of 0.01, and batch size 64. All LLM queries are conducted using GPT-4-turbo with default decoding parameters provided by the API. We acknowledge that GPT-4-turbo is a closed-source model subject to potential updates. In practice, our framework does not rely on model-specific properties, and can be readily extended to alternative LLMs with comparable capabilities, including open-source ones. Generated unknown-class labels are post-processed using an overlap-checking strategy to ensure that no generated label duplicates any known source class. If overlap occurs, the prompt is re-issued until a valid unknown label set is obtained.

CLIP is frozen, with ViT-B/16 as the image encoder and a masked self-attention Transformer \cite{vaswani2017attention} as the text encoder, performing only zero-shot inference. To account for variability in LLM outputs for unknown-category text, each task is executed three times using the same set of prompts, and the results reported in the tables of Section IV are averaged across runs. The scaling factor \(t\) in Eq.~(\ref{aSim}) is empirically set to 0.01. We fix \(K=5\) for neighborhood soft voting in Eq.~(\ref{eq:neigborVote}) and set \(\alpha=2\), assuming the number of unknown categories is twice that of known ones in the target domain. In our implementation, the neighborhood soft-voting adopts a direct nearest neighbor search, where cosine similarities between all target samples are computed and the top-\(K\) neighbors are selected by similarity ranking, without using approximate nearest neighbor (ANN) libraries \cite{johnson2019billion}. The factor \(\beta\) in Eq.~(\ref{eq3}) is set to 0.25 for Office-31 and DomainNet, and 0.75 for Office-Home and VisDA. The rationale for these choices is detailed in subsection~\ref{Experimental Analysis}. In the LSTD module, \(\delta=1.0\) is used, as justified in Fig.~\ref{fig5}. Following GLC \cite{qu2023upcycling}, the threshold \(w\) in Eq.~(\ref{eq13}) is set to 0.55. Due to the source-free setting and the absence of labeled target data, we adopt fixed hyperparameters rather than learning them end-to-end, and empirically show their robustness through systematic sensitivity analysis.

\subsection{Evaluation Metric and Compared Methods}
We evaluate our method across all possible adaptation shifts: OSDA, PDA, and OPDA. The details of the class splits are provided in Table~\ref{tab:label_split}. We use the evaluation metrics employed in previous works \cite{qu2024lead}. For PDA, we report classification accuracy. For OPDA, we use the H-score, which is the harmonic mean of accuracy on common data and accuracy on private data.
For OSDA, we primarily adopt H-score as the evaluation metric. However, for the compared method named Co-learn++ \cite{zhang2025source}, we report accuracy on Office-Home following its original paper, as H-score results are unavailable due to the absence of an official OSDA implementation. Methods designed to address category shifts generally do not perform well in the traditional closed-set domain adaptation (CLDA). To verify the versatility of LFM, we also conduct CLDA experiments on Office-31 and Office-Home with the accuracy as the metric. 

To ensure fair performance comparisons, we enforce two key principles. First, 
all compared methods are required to adopt the same ResNet-based backbone as ours for their target models. Although our approach leverages FMs during training, the final deployed target model retains the same ResNet-based architecture as traditional domain adaptation methods. Therefore, to ensure a fair comparison in terms of inference-time architecture, methods employing ViT-based extractors are excluded from comparison. Specifically, ``fairness'' here refers to maintaining the same target model architecture at deployment phase across all methods, while allowing differences in training-time auxiliary components.
Second, as a SF-UniDA method, we primarily compare against methods that explicitly claim to handle universal settings, even if some do not evaluate across all possible label shift scenarios. Except for CLDA, we do not include domain adaptation methods assuming a closed-set setting, as prior studies \cite{jing2023WDAN,jingNENO} have demonstrated their poor performance under label shifts. 

To strengthen the empirical comparison and isolate the contribution of our adaptation mechanism from the raw capability of foundation models, we include two CLIP-based baselines: CLIP Zero-Shot$^{\dagger}$ and LLM+CLIP. The CLIP Zero-Shot$^{\dagger}$ baseline evaluates CLIP's zero-shot capability in the SF-UniDA setting. Target samples are assigned to known classes based on text–image similarity (raw CLIP zero-shot), while unknown classes are identified using an entropy-based criterion on the normalized known-class probability distribution. This approach aligns with the unknown detection strategy in our method, ensuring the comparison focuses on CLIP's raw ability rather than unknown-class detection.
The LLM+CLIP baseline classifies target-domain samples by matching CLIP visual features with text embeddings of both source-known and LLM-generated unknown-class labels. It relies on fixed semantic alignment, without modeling label shift or performing target-adaptive refinement, making it a non-adaptive FM-based baseline for SF-UniDA.
For a fair comparison, all CLIP-based components in our method and the CLIP-related baselines use the same frozen CLIP configuration (ViT-B/16 image encoder), unless otherwise specified by the original implementation of the compared method.

\subsection{Experimental Results}

\begin{table*}[t]\renewcommand\arraystretch{1.2}
\centering
\scalebox{0.63}{
\begin{tabular}{lccc|ccccccccccccc|ccccccc|c}
\toprule
\multirow{2}[1]{*}{Methods} & \multirow{2}[1]{*}{SF} & \multirow{2}[1]{*}{U} & \multirow{2}[1]{*}{FM} & \multicolumn{13}{c|}{Office-Home}                                                           & \multicolumn{7}{c|}{Office-31}  &
\multicolumn{1}{c}{VisDA} \\
\cline{5-25} & & & & Ar2Cl & Ar2Pr & Ar2Re & Cl2Ar & Cl2Pr & Cl2Re & Pr2Ar & Pr2Cl & Pr2Re & Re2Ar & Re2Cl & Re2Pr & \textbf{\ Avg.\ }   & A2D & A2W & D2A & D2W & W2A  & W2D  & \textbf{Avg.}   & \textbf{S2R} \\
\midrule
CMU \cite{fu2020learning} & \xmark & \xmark & \xmark   & 55.0  & 57.0  & 59.0  & 59.3  & 58.2  & 60.6  & 59.2  & 51.3  & 61.2  & 61.9  & 53.5  & 55.3  & 57.6 & 52.6 & 55.7 & 76.5 & 75.9 & 65.8 & 64.7  & 65.2  & 54.2 \\
DANCE \cite{saito2020universal} & \xmark & \cmark & \xmark & 6.5   & 9.0   & 9.9   & 20.4  & 10.4  & 9.2   & 28.4  & 12.8  & 12.6  & 14.2  & 7.9   & 13.2  & 12.9 & 84.9 & 78.8 & 79.1 & 78.8 & 68.3 & 78.8 & 79.8  & 67.5 \\
DCC \cite{li2021domain} & \xmark  & \cmark  & \xmark  & 56.1  & 67.5  & 66.7  & 49.6  & 66.5  & 64.0  & 55.8  & 53.0  & 70.5  & 61.6  & 57.2  & 71.9  & 61.7 & 58.3 & 54.8 & 67.2 & 89.4 & 85.3 & 80.9 & 72.7  & 59.6 \\
OVANet \cite{saito2021ovanet} & \xmark & \xmark & \xmark   & 58.6  & 66.3  & 69.9  & 62.0  & 65.2  & 68.6  & 59.8  & 53.4  & 69.3  & 68.7  & 59.6  & 66.7  & 64.0 & 90.5 & 88.3 & 86.7 & \textbf{98.2} & 88.3 & \textbf{98.4}  & 91.7 & 66.1 \\
GATE \cite{chen2022geometric} & \xmark & \cmark & \xmark  & 63.8  & 70.5  & 75.8  & 66.4  & 67.9  & 71.7  & 67.3 & 61.5  & 76.0  & 70.4  & 61.8 & 75.1  & 69.0  & 88.4 & 86.5 & 84.2 & 95.0 & 86.1 & 96.7 & 89.5  & 70.8 \\
\midrule
Source-only &- &- &-    & 46.1  & 63.3  & 72.9  & 42.8  & 54.0  & 58.7  & 47.8  & {36.1} & 66.2  & 60.8  & 45.3  & 68.2  & 55.2  & 78.2 & 72.1 & 44.2 & 82.2 & 52.1 & 88.8 & 69.6      &29.1  \\
SHOT-O \cite{liang2020we} & \cmark & \xmark & \xmark & 37.7  & 41.8  & 48.4  & 56.4  & 39.8  & 40.9  & 60.0  & 41.5  & 49.7  & 61.8  & 41.4  & 43.6  & 46.9  & 80.2 & 71.6 & 64.3 & 93.1 & 64.0 & 91.8 & 77.5  & 28.1 \\
UMAD \cite{liang2021umad} & \cmark & \xmark &\xmark  & 59.2  & {71.8}  & 76.6  & 63.5  & 69.0  & 71.9  & 62.5  & 54.6  & 72.8  & 66.5  & 57.9  & 70.7  & 66.4 & 88.5 & 84.4 & 86.8 & 95.0 & 88.2 & 95.9  & 89.8  & 66.8 \\
GLC \cite{qu2023upcycling} & \cmark & \cmark &\xmark & 65.3    & 74.2     & 79.0       & 60.4       & 71.6      & 74.7     & 63.7     & 63.2      & 75.8     & 67.1      & 64.3     & 77.8 & {69.8}   & 82.6 & 74.6 & 92.6 & {96.0} & 91.8 & 96.1  & {89.0}      & 72.5  \\
LEAD \cite{qu2024lead} & \cmark & \cmark & \xmark & 60.7 & 70.8 & 76.5 & 61.0 & 68.6 & 70.8 & 65.5 & 59.8 & 74.2 & 64.8 & 57.7 & 75.8 & 67.2 & 84.9 & 85.1 & 90.2 & 94.8 & 90.3 & 96.5 & 90.3 & 74.2\\
{CLIP Zero-Shot$^{\dagger}$} &- &- &- & 55.3 & 36.6 & 40.7 & 53.2 & 36.6 & 40.7 & 53.2 & 55.3 & 40.7 & 53.2 & 55.3 & 36.6 & 46.5 & 80.4 & 69.3 & 61.0 & 69.3 & 61.0 & 80.4 & 70.2 & 66.0\\
{LLM+CLIP} &- &- &- & 65.6 & 82.4 & 81.0 & 77.5 & 82.4 & 81.0 & 77.5 & 65.6 & 81.0 & 77.5 & 65.6 & 82.4 & 76.6 & 92.7 & 86.0 & 90.7 & 86.0 & 90.7 & 92.7 & 89.8 & 83.2\\
SF-ODA w/ CLIP \cite{yu2025open} & \cmark & \xmark & \cmark & \textbf{76.5} & 79.3 & \textbf{85.5} & \textbf{82.3} & 78.6 & \textbf{84.8} & \textbf{82.2} & \textbf{76.5} & \textbf{84.6} & \textbf{82.1} & \textbf{76.8} & 78.9 & \textbf{80.7} & 93.3 & \textbf{89.3} & 91.9 & {92.6} & 90.8 & 93.5 & 91.9 & 83.8\\
DIFO$^{{\star}}$-C-B32 \cite{tang2024source} &\cmark &\xmark &\cmark &45.2 &56.8 &56.7 &50.8 &52.6 &53.5 &54.8 &45.2 &56.2 &53.5 &46.4 &55.7 &52.3 
 &66.8 & 66.2 &64.4 &74.1 &65.3 &74.5 &68.5  &38.4\\
Co-learn++$^{{\ddagger}}$
 \cite{zhang2025source} & \cmark & \cmark & \cmark & 54.9 & 77.6 & 78.4 & 60.4 & 72.7 & 75.9 & 56.0 & 51.0 & 77.2 & 64.0 & 58.8 & \textbf{83.6} & 67.5 & - & - & - & - & - & - & - & -\\
\hline
\rowcolor{Gray} LFM (Ours) & \cmark & \cmark & \cmark & {67.3}  &\textbf{83.2}  & {83.5}        & {76.6}        &\textbf{83.5}  & {83.4}        & {76.5}        &{67.4} &{83.4}  &{77.0} & {67.3}        & {83.4}        & {77.7} &\textbf{96.0} &{89.2} &\textbf{93.1}  &90.9   &\textbf{93.1}  &97.4  &\textbf{93.3}  &\textbf{86.9} \\
\hline
\end{tabular}
}
\caption{H-score (\%) comparison under the OSDA setting on Office-Home, Office-31, and VisDA. $^\ddagger$ Results of Co-learn++$^{{\ddagger}}$ on Office-Home under OSDA are reported using accuracy, as provided in the original paper. Since its official implementation does not support OSDA settings, H-score results cannot be obtained in a reproducible manner. $^{\dagger}$ CLIP Zero-Shot$^{\dagger}$ denotes an enhanced zero-shot baseline, where CLIP is equipped with the same entropy-based unknown detection strategy as our method, since vanilla CLIP zero-shot inference only identifies source-known classes.
$^{\star}$ Results of DIFO$^{{\star}}$-C-B32 are reproduced or newly extended by us based on the authors’ publicly released implementations, rather than directly reported in the original papers. All reproduced results follow the same evaluation protocol and metrics as our method for fair comparison.
}
\label{tab:osda}
\end{table*}%

\begin{table*}[t]\renewcommand\arraystretch{1.2}
\centering
\scalebox{0.636}{
\begin{tabular}{lccc|ccccccccccccc|ccccccc|c}
\toprule
\multirow{2}[1]{*}{Methods} & \multirow{2}[1]{*}{SF} & \multirow{2}[1]{*}{U} & \multirow{2}[1]{*}{FM} & \multicolumn{13}{c|}{Office-Home}                                                           & \multicolumn{7}{c|}{Office-31}  &
\multicolumn{1}{c}{VisDA} \\
\cline{5-25} & & & & Ar2Cl & Ar2Pr & Ar2Re & Cl2Ar & Cl2Pr & Cl2Re & Pr2Ar & Pr2Cl & Pr2Re & Re2Ar & Re2Cl & Re2Pr & \textbf{\ Avg.\ }   & A2D & A2W & D2A & D2W & W2A  & W2D  & \textbf{Avg.}   & \textbf{S2R} \\
\midrule
CMU \cite{fu2020learning} & \xmark & \xmark & \xmark & 50.9  & 74.2  & 78.4  & 62.2  & 64.1  & 72.5  & 63.5  & 47.9  & 78.3  & 72.4  & 54.7  & 78.9  & 66.5 & 84.1 & 84.2  &69.2 & 97.2 & 66.8 & 98.8 & 83.4  &65.5\\
DANCE \cite{saito2020universal} & \xmark & \cmark & \xmark & 53.6  & 73.2  & 84.9  & 70.8  & 67.3  & 82.6  & 70.0  & 50.9  & 84.8  & 77.0  & 55.9  & 81.8  & 71.1 & 77.1 & 71.2 & 83.7 & 94.6 & 92.6 & 96.8 &86.0 & 73.7 \\
DCC \cite{li2021domain}  & \xmark & \cmark & \xmark & 54.2  & 47.5  & 57.5  & 83.8  & 71.6  & 86.2  & 63.7  & 65.0 & 75.2  & 85.5 & 78.2  & 82.6  & 70.9 & 87.3 & 81.3 & 95.4 & \textbf{100.0} & 95.5 & \textbf{100.0} & 93.3 & 72.4\\
OVANet \cite{saito2021ovanet} & \xmark & \xmark & \xmark & 34.1  & 54.6  & 72.1  & 42.4  & 47.3  & 55.9  & 38.2  & 26.2  & 61.7  & 56.7  & 35.8  & 68.9  & 49.5 & 69.4 & 61.7 & 61.4 & 90.2 & 66.4 & 98.7 & 74.6 & 34.3 \\
GATE \cite{chen2022geometric} & \xmark & \cmark & \xmark & 55.8  & 75.9  & 85.3  & 73.6  & 70.2  & 83.0  & 72.1  & 59.5  & 84.7  & 79.6  & 63.9  & 83.8  & 74.0 & 89.5 & 86.2 & 93.5 & \textbf{100.0} & 94.4 & 98.6 & 93.7 & 75.6 \\
\midrule
Source-only &- &- &- & 45.9  & 69.2  & 81.1  & 55.7  & 61.2  & 64.8  & 60.7  & 41.1  & 75.8  & 70.5  & 49.9  & 78.4  & 62.9   & 90.4 & 79.3 & 79.3 & 95.9 & 84.3 & 98.1 & 87.8 & 42.8  \\
SHOT-P \cite{liang2020we} & \cmark & \xmark & \xmark & 64.7  & 85.1  & 90.1  & 75.1  & 73.9 & 84.2  & 76.4  & 64.1  & 90.3  & 80.7  & 63.3  & 85.5  & 77.8 & 89.8 & 84.4 & 92.2 & 96.6 & 92.2 & {99.4} & 92.4 & 74.2\\
UMAD \cite{liang2021umad} & \cmark & \xmark &\xmark & 51.2     & 66.5     & 79.2     & 63.1     & 62.9     & 68.2     & 63.3     & 56.4     & 75.9     & 74.5     & 55.9     & 78.3     & 66.3  & 85.4 & 85.1 & 83.5 & 97.6 & 86.2 & {99.4} & 89.5 & 68.5 \\
GLC \cite{qu2023upcycling} & \cmark & \cmark  & \xmark & 55.9      & 79.0      & 87.5      & 72.5      & 71.8      & 82.7      & 74.9      & 41.7      & 82.4      & 77.3      & 60.4      & 84.3      & 72.5 & 89.8 & 89.8 & 92.8 & 96.6 & \textbf{96.1} & {99.4} & 94.1 & 76.2\\
LEAD \cite{qu2024lead} & \cmark &\cmark & \xmark & 58.2 & 83.1 & 87.0 & 70.5 & 75.4 & 83.3 & 73.7 & 50.4 & 83.7 & 78.3 & 58.7 & 83.2 & 73.8 & 89.8 & 93.9 & 95.6 & {98.6} & 96.0 & {99.4} & 95.5  & 75.3\\
{CLIP Zero-Shot$^{\dagger}$} &- &- &- &72.4 &87.3 &90.0 &83.5 &87.3 &90.0 &83.5 &72.4 &90.0 &83.5 &72.4 &87.3 &83.3 &88.5 &87.5 &87.9 &87.5 &87.9 &88.5 &88.0  &87.1 \\
{LLM+CLIP} &- &- &- &69.4 &85.9 &88.5 &79.7 &85.9 &88.5 &79.7 &69.4 &88.5 &79.7 & 69.4 &85.9 &80.9 &84.1 &81.2 &84.0 &81.2 &84.0 &84.1 &83.1 &72.4 \\
DIFO$^{{\star}}$-C-B32 \cite{tang2024source} & \cmark & \xmark & \cmark & {70.2} & \textbf{91.7} & {91.5} & {87.8} & \textbf{92.6} & {92.9} & {87.3} & {70.7} & {92.9} & {88.5} & {69.6} & {91.5} & {85.6} &90.1 &91.2 &80.3 &96.6  &79.6 &99.6 &89.6 & 88.1 
\\
Co-learn++ \cite{zhang2025source} & \cmark &\cmark & \cmark & 68.0 & 86.6 & 91.7 & 77.7 & 73.1 & 89.3 & 79.2 & 68.3 & 91.2 & 83.3 & 68.2 & 87.5 & 80.3 & - & - & - & - & - & - & - & - \\         
\hline
\rowcolor{Gray} LFM (Ours) & \cmark &\cmark &\cmark &\textbf{80.9}      &\textbf{91.7}  &\textbf{94.4}  &\textbf{89.6}  &{92.3} &\textbf{94.9}  &\textbf{89.4}  &\textbf{80.8}  &\textbf{94.4}  &\textbf{90.1}  &\textbf{80.5}  &\textbf{92.3}  &\textbf{89.3} &\textbf{96.2}   &\textbf{96.3}  &\textbf{96.0}  &97.6   &96.0   &{99.4}  &\textbf{96.9}  &\textbf{93.4} \\
\hline
\end{tabular}
}
\caption{Accuracy comparison (\%) under the PDA setting on Office-Home, Office-31, and VisDA.
}
\label{tab:pda}%
\end{table*}%

\begin{table*}[!htp]\renewcommand\arraystretch{1.2}
\centering
\scalebox{0.75}{
    \begin{tabular}{lccc|ccccccl|ccccccc|c}
        \toprule
        \multirow{2}[1]{*}{Methods} & \multirow{2}[1]{*}{SF} & \multirow{2}[1]{*}{U} & \multirow{2}[1]{*}{FM} & \multicolumn{7}{c|}{Office-31} & \multicolumn{7}{c|}{DomainNet} & \multicolumn{1}{c}{VisDA}\\
        \cline{5-19}  &  &    &  & A2D   & A2W   & D2A   & D2W   & W2A   & W2D   & \textbf{Avg.}  & {P2R} & {P2S} & {R2P} & {R2S} & {S2P} & {S2R} & \textbf{Avg.} &  \textbf{S2R}\\
        \hline
        UAN \cite{you2019universal} & \xmark & \xmark & \xmark &86.5 &85.6 &85.5 &94.8 &85.1 &\textbf{98.0}  &89.2  & - & - & - & - & - & - & - &60.8 \\
        CMU \cite{fu2020learning} & \xmark & \xmark & \xmark & 68.1  & 67.3  & 71.4  & 79.3  & 72.2  & 80.4  & 73.1    & {50.8} & {{45.1}} & {52.2} & {45.6} & {44.8} & {51.0} & {48.3} & 32.9\\
        DANCE \cite{saito2020universal} & \xmark & \cmark & \xmark & 78.6  & 71.5  & 79.9  & 91.4  & 72.2  & 87.9  & 80.3    & {21.0}  & {37.0} & {47.3} & {46.7} & {27.7}  & {21.0} & {33.5} & 42.8\\
        DCC \cite{li2021domain} & \xmark & \cmark & \xmark & {88.5}  & 78.5  & 70.2  & 79.3  & 75.9  & 88.6  & 80.2    & {56.9} & {43.7} & {50.3} & {43.3} & {44.9} & {56.2} & {49.2} & 43.0\\
        OVANet \cite{saito2021ovanet} & \xmark & \xmark & \xmark & 85.8  & 79.4  & 80.1  & {95.4}  & 84.0  &{94.3}  & 86.5    & {56.0} & {47.1} & {51.7} & {44.9} & {47.4}  & {57.2} & {50.7} & 53.1\\
        GATE \cite{chen2022geometric} & \xmark & \cmark & \xmark & {87.7}  & {81.6} & 84.2  &94.8  & 83.4  & 94.1  & 87.6    & 57.4 & 48.7 & 52.8  & 47.6 & 49.5 & {56.3} & 52.1 & 56.4\\
        UniOT \cite{chang2022unified} & \xmark &\xmark &\xmark & 87.0 & \textbf{88.5} & 88.4 & \textbf{98.8} & 87.6 & 96.6 & \textbf{91.1} & 59.3 & 51.8 & 47.8 & 48.3 & 46.8 & 58.3 & 52.0 & 57.3\\
        \midrule
        Source-only &- & - & -& 70.9  & 63.2  & 39.6  & 77.3  & 52.2  & 86.4  & 64.9    & 57.3      & 38.2      & 47.8      & 38.4      & 32.2      & 48.2      & 43.7 & 25.7\\
        SHOT-O \cite{liang2020we} & \cmark & \xmark & \xmark & 73.5  & 67.2  & 59.3  & 88.3  & 77.1  & 84.4  & 75.0    & {35.0} & {30.8} & {37.2} & {28.3} & {31.9} & {32.2} & {32.6} & 44.0\\
        UMAD \cite{liang2021umad} & \cmark & \xmark & \xmark & 79.1  & 77.4  & 87.4  & 90.7  & \textbf{90.4}  & {97.2}  & 87.0   & 59.0 & {44.3} &  {50.1} & {42.1} & {32.0} & {55.3} & {47.1} & 58.3 \\
        USFDA \cite{kundu2020universal} & \cmark & \xmark & \xmark & 88.5  & {85.6}  & 87.5  & 95.2  & {86.6}  & {97.8}  &  90.2  & - & - & - & - & - & - & - & 63.9 \\
        GLC \cite{qu2023upcycling} & \cmark & \cmark & \xmark & {81.5} & 84.5 & \textbf{89.8} & 90.4 &88.4 & 92.3  & {87.8} & 63.3  & 50.5  & 54.9    & 50.9   & 49.6    & 61.3     & {55.1} & 73.1\\
        LEAD \cite{qu2024lead} & \cmark &\cmark & \xmark & 85.4 & {85.0} & 86.3 & {90.9} & 86.2 & 93.1 & {87.8} & 59.9 & 46.1 & 51.3 & 45.0 & 45.9 & 56.3 & 50.8 & 76.6 \\
        {CLIP Zero-Shot$^{\dagger}$}  & - & - & - & 60.1 & 57.3  & 57.0  & 57.3    & 57.0  & 60.1  & 58.1 & 54.3   & 56.1  & 60.3  & 56.1  & 60.3  & 54.3  & 56.9  & 61.5 \\
        {LLM+CLIP}  & - & - & -        & 80.8 & 75.1  & 80.9  & 75.1    & 80.9  & 80.8  & 78.9 
 & 76.0 & \textbf{63.6}  & \textbf{67.2}  & 63.6 & \textbf{67.2} & 76.0 & \textbf{68.9}  & 75.8 \\
        DIFO$^{{\star}}$-C-B32 \cite{tang2024source} & \cmark & \xmark & \cmark & 48.7  &  46.1   &  50.3   &  57.7   &  47.2   &  55.2   &  50.9 
              & 47.4  & 41.0    & 45.4    & 43.6    & 38.9    & 47.7    & 44.0 & 41.3 \\
        \hline
        \rowcolor{Gray} LFM (Ours) & \cmark &\cmark & \cmark &\textbf{89.4} &81.9 &88.0   &83.9 & 88.1 & 94.0 & 87.6 &\textbf{76.4} & 63.2 &63.6  &\textbf{63.7}  &63.1  &\textbf{76.3}  &67.7 &\textbf{86.2} \\
        \hline
    \end{tabular}
}
\caption{H-score (\%) comparison under the OPDA setting on Office-31, DomainNet, and VisDA.  
}
\label{tab:opda_rest}
\end{table*}%

\begin{table*}[!htp]\renewcommand\arraystretch{1.2}
\centering
\scalebox{0.75}{
    \begin{tabular}{lccc|ccccccccccccl}
        \toprule
        \multirow{2}[1]{*}{Methods} & \multirow{2}[1]{*}{SF} & \multirow{2}[1]{*}{U} & \multirow{2}[1]{*}{FM} & \multicolumn{13}{c}{Office-Home}  \\
        \cline{5-17} & & & & Ar2Cl & Ar2Pr & Ar2Re & Cl2Ar & Cl2Pr & Cl2Re & Pr2Ar & Pr2Cl & Pr2Re & Re2Ar & Re2Cl & Re2Pr & \textbf{Avg.} \\
        \hline
        UAN \cite{you2019universal} & \xmark & \xmark & \xmark &63.0 &82.8 &87.9 &76.9 &78.7 &85.4 &78.2 &58.6 &86.8 &83.4 &63.2 &79.4 &77.0   \\
        CMU \cite{fu2020learning} & \xmark & \xmark & \xmark &  56.0  & 56.9  & 59.2  & 67.0  & 64.3  & 67.8  & 54.7  & 51.1  & 66.4  & 68.2  & 57.9  & 69.7  & 61.6  \\
        DANCE \cite{saito2020universal} & \xmark & \cmark & \xmark & 61.0  & 60.4  & 64.9  & 65.7  & 58.8  & 61.8  & 73.1  & 61.2  & 66.6  & 67.7  & 62.4  & 63.7  & 63.9  \\
        DCC \cite{li2021domain} & \xmark  & \cmark  & \xmark & 58.0  & 54.1  & 58.0  & 74.6  & 70.6  & 77.5  & 64.3  & {73.6}  & 74.9  & {81.0} & {75.1} & 80.4  & 70.2  \\
        OVANet \cite{saito2021ovanet}  & \xmark & \xmark &\xmark & 62.8  & 75.6  & 78.6  & 70.7  & 68.8  & 75.0  & 71.3  & 58.6  & 80.5  & 76.1  & 64.1  & 78.9  & 71.8  \\
        GATE \cite{chen2022geometric} & \xmark &\cmark &\xmark & 63.8  & {75.9}  & {81.4}  & {74.0}  & {72.1}  & 79.8  & 74.7  & {70.3}  & {82.7} & 79.1  & 71.5  & {81.7}  & 75.6  \\
        UniOT \cite{chang2022unified} & \xmark &\xmark &\xmark & 67.3 & 80.5 & 86.0 & 73.5 & 77.3 & 84.3 & 75.5 & 63.3 & 86.0 & 77.8 & 65.4 & 81.9 & {76.6} \\     
        \midrule
        Source-only &- &- &- & 47.3  & 71.6  & 81.9  & 51.5  & 57.2  & 69.4  & 56.0  & 40.3  & 76.6  & 61.4  & 44.2  & 73.5  & 60.9  \\
        SHOT-O \cite{liang2020we} & \cmark & \xmark & \xmark & 32.9  & 29.5  & 39.6  & 56.8  & 30.1  & 41.1  & 54.9  & 35.4  & 42.3  & 58.5  & 33.5  & 33.3  & 40.7  \\
        UMAD \cite{liang2021umad} & \cmark  & \xmark  & \xmark & 61.1  & 76.3  & 82.7  & 70.7  & 67.7  & 75.7  & 64.4  & {55.7}  & 76.3  & 73.2  & {60.4}  & 77.2  & 70.1  \\
        USFDA \cite{kundu2020universal} & \cmark  & \xmark  & \xmark &63.4 &83.3 &89.4 &71.0 &72.3 &86.1 &78.5 &60.2 &87.4 &81.6 &63.2 &\textbf{88.2} &77.1  \\
        GLC \cite{qu2023upcycling} & \cmark  & \cmark  &\xmark & 64.3     & 78.2     & \textbf{89.8}     & 63.1      & 81.7  & 89.1     &{77.6}      & 54.2      & {88.9}    & 80.7    & 54.2      & 85.9     & {75.6} \\
        LEAD \cite{qu2024lead} & \cmark &\cmark & \xmark  & 62.7 & 78.1 & 86.4 & 70.6 & 76.3 & 83.4 & 75.3 & 60.6 & 86.2 & 75.4 & 60.7 & 83.7 & 75.0 \\
        {CLIP Zero-Shot$^{\dagger}$} & - & - & - & 64.7 & 43.1 & 53.5 & 67.8 & 43.1 & 53.5 & 67.8 & 64.7 & 53.5 & 67.8 & 64.7 & 43.1 & 57.3 \\
        {LLM+CLIP}        & - & - & - & 57.9 & 70.0 & 69.0 & 75.1 & 70.0 & 69.0 & 75.1 & 57.9 & 69.0 & 75.1 & 57.9 & 70.0 & 68.0 \\
        Co-learn++ \cite{zhang2025source} & \cmark &\cmark & \cmark  & 63.4 & 86.6 & 87.3 & 74.1 & 77.9 & 83.3 & 81.5 & 66.6 & 89.3 & 82.4 & 70.0 & 86.2 & 79.0\\
        DIFO$^{{\star}}$-C-B32 \cite{tang2024source} & \cmark &\xmark & \cmark &39.9 &47.5 &50.7 &39.6 &39.9 &43.3 &41.6 &37.3 &47.7 &42.7 &41.0 &43.3 & 42.9\\
        \hline
        \rowcolor{Gray} LFM (Ours) & \cmark &\cmark & \cmark  &\textbf{78.1}    &\textbf{87.2}  &89.6   &\textbf{86.8}  &\textbf{87.1}  &\textbf{89.5}   &\textbf{86.7}  &\textbf{77.9}  &\textbf{89.7}   &\textbf{86.8}  &\textbf{77.4}  &{87.2}  &\textbf{85.3}\\
        \hline
    \end{tabular}
}
\caption{H-score (\%) comparison under the OPDA setting on Office-Home.}
\label{tab:opda_officehome}
\end{table*}

\begin{table*}[!tph]\renewcommand\arraystretch{1.2}
    \centering
    \scalebox{0.70}{
    {}\begin{tabular}{lccc|ccccccccccccc|c}
        \toprule
        \multirow{2}[1]{*}{Methods} & \multirow{2}[1]{*}{SF} & \multirow{2}[1]{*}{U/C} & \multirow{2}[1]{*}{FM} & \multicolumn{13}{c|}{Office-Home} & \multicolumn{1}{c}{Office-31}  \\
        \cline{5-18} & & & & Ar2Cl & Ar2Pr & Ar2Re & Cl2Ar & Cl2Pr & Cl2Re & Pr2Ar & Pr2Cl & Pr2Re & Re2Ar & Re2Cl & Re2Pr & \textbf{Avg.}  & \textbf{Avg.}  \\
        \midrule
        CDAN \cite{long2018conditional}  & \xmark & \xmark/\cmark & \xmark & 49.0 &69.3 &74.5 &54.4 &66.0 &68.4 &55.6 &48.3 &75.9 &68.4 &55.4 &80.5 &63.8 &86.6\\
        MDD \cite{zhang2019bridging}  & \xmark & \xmark/\cmark & \xmark & 54.9 &73.7 &77.8 &60.0 &71.4 &71.8 &61.2 &53.6 &78.1 &72.5 &60.2 &82.3 &68.1 &{88.9}\\
        UAN \cite{you2019universal} & \xmark & \xmark/\xmark & \xmark & 45.0 &63.6 &71.2 &51.4 &58.2 &63.2 &52.6 &40.9 &71.0 &63.3 &48.2 &75.4 &58.7 &84.4\\
        CMU \cite{fu2020learning} & \xmark & \xmark/\xmark & \xmark & 42.8  & 65.6  & 74.3  & 58.1  & 63.1  & 67.4  & 54.2  & 41.2  & 73.8  & 66.9  & 48.0  & 78.7  & 61.2 & 79.9\\
        DANCE \cite{saito2020universal} & \xmark & \cmark/\xmark & \xmark & 54.3 &75.9 &78.4 &64.8 &72.1 &73.4 &63.2 &53.0 &79.4 &73.0 &58.2 &82.9 &69.1 &85.5 \\
        DCC \cite{li2021domain} & \xmark & \cmark/\xmark & \xmark & 35.4 &61.4 &75.2 &45.7 &59.1 &62.7 &43.9 &30.9 &70.2 &57.8 &41.0 &77.9 &55.1 &87.4\\
        OVANet \cite{saito2021ovanet} & \xmark & \xmark/\xmark & \xmark & 34.5 &55.8 &67.1 &40.9 &52.8 &56.9 &35.4 &26.2 &61.8 &53.8 &35.4 &70.8 &49.3 &70.4 \\
        \midrule
        Source-only & - & - & - & 44.8 &67.4 &74.2 &53.0 &63.3 &65.1 &53.7 &40.5 &73.5 &65.6 &46.3 &78.3 &60.5 &78.8  \\
        SHOT \cite{liang2020we} & \cmark & \xmark/\cmark & \xmark &57.1 &78.1 &81.5 &68.0 &78.2 &78.1 &67.4 &54.9 &82.2 &73.3 &58.8 &84.3 &71.8 &88.6 \\
        {SHOT \cite{liang2020we} w/ DCPL \cite{eccv2024DCPL}} & \cmark & \xmark/\cmark & \xmark &59.9 &84.3 &87.8 &76.8 &85.8 &86.6 &74.8 &58.6 &87.4 &77.9 &61.1 &89.0 &77.5 &-\\
        {HRD \cite{xing2024hierarchical}} & \cmark & \xmark/\cmark & \xmark &61.2 &82.6 &84.5 &70.8 &81.4 &82.5 &69.9 &59.7 &84.0 &75.2 &63.6 &87.4 &75.2 &90.6 \\
        UMAD \cite{liang2021umad} & \cmark & \xmark/\xmark & \xmark & 48.0 &65.1 &73.0 &58.6 &65.3 &67.9 &58.2 &47.3 &74.0 &69.4 &53.0 &77.8 &63.1 &81.7 \\
        GLC \cite{qu2023upcycling} & \cmark & \cmark/\xmark & \xmark & 51.2 &76.0 &79.9 &65.4 &78.6 &78.7 &65.6 &54.1 &81.6 &70.9 &58.4 &84.2 &70.4 &88.1\\
        {CLIP Zero-Shot$^{\dagger}$} & - & - & - &66.7 &86.6 &86.3  &78.4  &86.6  &86.3 &78.4   &66.7  &86.3  &78.4   &66.7 &86.6  &79.5  &79.8 \\ 
        {LLM+CLIP}        & - & - & - &60.8 &86.5 &87.1 & 78.2 & 86.5 & 87.1 & 78.2 & 60.8 & 87.1 & 78.2 & 60.8 & 86.5 & 78.2  &74.2 \\
        DIFO-C-B32 \cite{tang2024source} & \cmark & \xmark/\cmark & \cmark
        &\textbf{70.6} &\textbf{90.6} &88.8 &82.5 &\textbf{90.6} &88.8 &80.9 &\textbf{70.1} &88.9 &83.4 &\textbf{70.5} &\textbf{91.2} &83.1 
        &\textbf{92.5}\\
        \hline
        \rowcolor{Gray} LFM (Ours) & \cmark & \cmark/\xmark & \cmark & {66.6}  &{90.4}  &\textbf{90.8}  &\textbf{84.2}  &\textbf{90.6}  &\textbf{91.0}  &\textbf{84.1}  &{67.9}  &\textbf{91.0}  &\textbf{85.1}  &{67.7}  &{90.7}  &\textbf{83.3} &{88.9} \\
        \hline
    \end{tabular}
    }
    \caption{Accuracy (\%) comparison under the CLDA setting on Office-Home and Office-31.}
    \vspace{-0.1in}
    \label{tab:clda}
\end{table*}

Experimental results are shown in Tables~\ref{tab:osda} -- \ref{tab:clda}, with the best results highlighted in \textbf{bold}. Based on their reliance on source data, the methods are categorized into two groups: source-free (``\cmark'' in the SF column) and those requiring source data. The FM column indicates whether a method leverages foundation models (``\cmark'' for yes). ``U'', short for universal, denotes whether the method is applicable for all potential label-shifts or not. The U/C column in Table~\ref{tab:clda} specifies whether each method was originally designed to support UniDA and/or CLDA. The results of the competitors are from the paper of LAED, GLC, or their original papers.

\noindent\textbf{Results for OSDA Scenarios.} 
The OSDA experiments are conducted on Office-Home, Office-31, and VisDA. As shown in Table~\ref{tab:osda}, LFM achieves the second-highest average H-score (Avg.) on Office-Home, following SF-ODA w/ CLIP \cite{yu2025open}, which is specifically designed for OSDA with CLIP. The LLM+CLIP baseline ranks third and outperforms two other CLIP-based methods, DIFO-C-B32 \cite{tang2024source} and Co-learn++ \cite{zhang2025source}, demonstrating the effectiveness of enhancing CLIP with LLM-generated semantic knowledge. 
In contrast, the CLIP Zero-Shot$^{\dagger}$ baseline performs poorly and even underperforms the Source-only baseline. This result suggests that, in the presence of unknown categories in the target domain, the rich semantic knowledge encoded in CLIP alone is insufficient to guarantee satisfactory performance without explicit target-adaptive mechanisms. 
When evaluated using H-score, DIFO-C-B32 \cite{tang2024source} performs worse than the source-only baseline under the OSDA setting of Office-Home and Office-31, suggesting limited effectiveness in jointly handling source-known classification and target-unknown recognition. Notably, all best task-wise results on Office-Home are achieved by CLIP-enhanced methods, underscoring the effectiveness of CLIP in OSDA. On Office-31, LFM achieves the best performance on four out of six tasks and the highest average H-score. 
For VisDA, LFM significantly outperforms all baselines. These results support the effectiveness of LFM in OSDA settings.

\noindent\textbf{Results for PDA Scenarios.} As shown in Table~\ref{tab:pda}, LFM achieves the highest average accuracy and delivers the best performance on nearly all tasks under the PDA setting across three benchmarks. Compared with previous methods, LFM demonstrates significant improvements. Specifically, LFM outperforms LEAD \cite{qu2024lead}, a state-of-the-art source-free universal method, by 15.5\%, 1.4\%, and 18.1\% in average accuracy on Office-Home, Office-31, and VisDA. There are in fact no unknown classes in the target domain under the PDA label shift. However, LLM+CLIP incorrectly assigns some known-class samples to the unknown class hypothesized by the LLM, as it does not explicitly model the label shift type. As a result, on all PDA tasks across the three benchmarks, the LLM+CLIP baseline underperforms CLIP Zero-Shot$^{\dagger}$. Moreover, among the methods that leverage FMs, namely CLIP Zero-Shot$^{\dagger}$, LLM+CLIP, Co-learn++, DIFO-C-B32, and ours, all surpass those without FMs by a notable margin on Office-31, underscoring the effectiveness of FMs in mitigating the challenges posed by source-private categories.

\noindent\textbf{Results for OPDA Scenarios.} We evaluate the effectiveness of LFM under the OPDA setting across four benchmarks, as shown in Table~\ref{tab:opda_rest} and Table~\ref{tab:opda_officehome}. Compared to methods that do not incorporate FMs, LFM demonstrates notable performance advantages on DomainNet and VisDA. Specifically, our method achieves Avg. improvements of 16.9\% and 9.6\%, respectively, over LEAD. Although the highest Avg. is achieved by UniOT \cite{chen2022geometric} on Office-31, it relies on access to source data, making it not directly comparable to source-free approaches. Moreover, LFM performs competitively with other state-of-the-art source-free methods, such as GLC \cite{qu2023upcycling} and LEAD \cite{qu2024lead}. On Office-Home, LFM achieves the highest Avg. and H-score scores on 10 out of 12 tasks, highlighting its robustness across diverse situations. Compared with Co-learn++ \cite{zhang2025source}, which also employs CLIP, LFM achieves a 6.3\% improvement in Avg., further demonstrating the advantage of our approach in exploiting the knowledge of FMs. DIFO has not previously been evaluated under the OPDA setting. Using the authors' publicly released implementation, we evaluate its OPDA performance on four benchmarks. Except on VisDA, DIFO either underperforms or shows similar performance compared to the source-only baseline in terms of average H-score across the remaining benchmarks. Combined with its degraded performance under OSDA, this suggests that DIFO may struggle with handling target-unknown classes, which are crucial in both OSDA and OPDA settings.

Regarding the two CLIP-based baselines, LLM+CLIP consistently outperforms CLIP Zero-Shot$^{\dagger}$ across all tasks on all benchmarks. Consistent with their comparison under the OSDA setting, this result further implies the effectiveness of the unknown-class text labels envisioned by the LLM in facilitating the recognition of unknown target categories. On DomainNet, LLM+CLIP achieves performance comparable to ours. We attribute this to the substantially larger data volume of DomainNet, which amplifies the benefit of the rich semantic knowledge embedded in FMs. Nevertheless, on the other three benchmarks, our method consistently outperforms LLM+CLIP, highlighting the necessity of our proposed target-adaptive refinement mechanism.    

\noindent\textbf{Results for CLDA Scenarios.}
Table~\ref{tab:clda} shows the results under the CLDA setting. Although LFM is not tailored for CLDA, it achieves competitive performance, with average accuracies of 83.3\% on Office-Home and 88.9\% on Office-31. This exceeds GLC \cite{qu2023upcycling}, a state-of-the-art SF-UniDA method without FMs, by 12.9\% and 0.8\%, respectively, confirming the benefit of leveraging the knowledge embedded in FMs. While DIFO-C-B32 \cite{tang2024source} achieves the best results on Office-31, LFM surpasses it on half the tasks and achieves higher average accuracy on the more challenging Office-Home benchmark. Notably, LFM attains these results without being specifically designed for CLDA, underscoring its versatility and robustness. Although CLIP Zero-Shot$^{\dagger}$ and LLM+CLIP do not match the performance of our method, both baselines consistently outperform other non–FM-based methods on Office-Home, indicating the effectiveness of CLIP zero-shot generalization. Consistent with the observations under PDA, where no unknown-class samples exist in the target domain, LLM+CLIP performs worse than CLIP Zero-Shot$^{\dagger}$, as it tends to assign some known-class target samples to unknown categories envisioned by the LLM.

\subsection{Experimental Analysis} \label{Experimental Analysis}
Before presenting detailed experimental analysis, we first clarify the perspective underlying the proposed framework. Rather than viewing our method as a collection of independent heuristics, we design it via formulating SF-UniDA as a structured label-shift reasoning problem. Accordingly, the proposed pipeline decomposes the problem into three logically connected stages: detecting the presence of unknown classes, separating known and unknown samples, and refining pseudo-labels with confidence and consensus-aware regularization.

The $cv$-based criterion in the LSTD module is designed as a robust indicator of distributional asymmetry between high- and low-score regions, which serves as a signature of unknown-class presence.
By focusing on the extremes of the score distribution, this statistic suppresses ambiguous mid-confidence samples and enables stable shift-type inference.
Based on the inferred shift type, unknown-class separation is formulated as a binary density estimation problem, where a two-component GMM represents a minimal and well-matched modeling assumption for distinguishing known and unknown classes.
Finally, the PLR module acts as an error-control mechanism that mitigates error propagation by enforcing consistency between model confidence and semantic consensus.

The following experimental analyses are organized to empirically validate the necessity and effectiveness of each component in this reasoning pipeline.

\begin{table}[!tp]
\renewcommand\arraystretch{1.2}
\centering
\scalebox{0.75}{
\begin{tabular}{l|ccc|ccc|ccc}
    \toprule
 \multirow{2}[1]{*}{}& \multicolumn{3}{c|}{Office-31} & \multicolumn{3}{c|}{Office-Home} & \multicolumn{3}{c}{VisDA}    \\
 \cline{2-10} & OPDA & OSDA & PDA & OPDA & OSDA & PDA & OPDA & OSDA & PDA\\ 
    \midrule
    w/o PLR  &85.6     &93.0   &94.9   &84.3   &77.5   &86.5   &82.8   &86.6   &91.8 
 \\
    LFM &87.6   &93.3   &96.9   &84.5   &77.7   &89.3   &83.4   &86.9   &93.4  \\
    \bottomrule
\end{tabular}
}
\caption{Ablation studies on the refinement module.}
\label{tab:ab}
\end{table}

\begin{table}[!tp]
\renewcommand\arraystretch{1.2}
\centering
\scalebox{0.75}{
\begin{tabular}{l|cccc|cccc|cccc}
\toprule
\multirow{2}{*}{Percentage} 
& \multicolumn{4}{c|}{OSDA} 
& \multicolumn{4}{c|}{OPDA} 
& \multicolumn{4}{c}{PDA} \\
\cline{2-13}
& Ar & Cl & Pr & Rw & Ar & Cl & Pr & Rw & Ar & Cl & Pr & Rw \\
\midrule
10\% & 1.5 & 1.5 & 1.6 & 1.6 & 1.6 & 1.6 & 1.6 & 1.6 & 0.9 & 0.9 & 0.6 & 0.6 \\
20\% & 1.4 & 1.3 & 1.5 & 1.5 & 1.5 & 1.5 & 1.6 & 1.5 & 0.8 & 0.8 & 0.5 & 0.5 \\
30\% & 1.3 & 1.4 & 1.3 & 1.3 & 1.7 & 1.6 & 1.5 & 1.6 & 0.6 & 0.4 & 0.3 & 0.6 \\
40\% & 1.2 & 1.1 & 1.2 & 1.1 & 1.7 & 1.7 & 1.3 & 1.5 & 0.6 & 0.6 & 0.4 & 0.3 \\
50\% & 1.1 & 1.1 & 1.0 & 1.0 & 1.7 & 1.8 & 1.7 & 1.5 & 0.5 & 0.5 & 0.3 & 0.3 \\
\bottomrule
\end{tabular}
}
\caption{Sensitivity analysis of $cv$ with respect to the percentage of selected scores on Office-Home.}
\label{tab:ScorePercentage}
\end{table}

\noindent\textbf{Ablation Study.}
To validate the effectiveness of the proposed pseudo-label refinement (PLR) module described in Section~\ref{PLR}, we conduct an ablation study by removing the PLR module from the LFM framework. In this variant, denoted as w/o PLR in Table~\ref{tab:ab}, the pseudo-labels predicted by CLIP are directly used without refinement. Across all types of label shifts and benchmarks, this variant consistently underperforms compared to the full LFM framework. Specifically, while the PLR module yields modest improvements under the OPDA and OSDA settings, it brings significantly larger gains under the PDA scenario. This can be attributed to the fact that all target categories are present in the source domain under PDA, making the task-specific knowledge embedded in the target model transferred from the source model crucial for accurate classification. The PLR module is specifically designed to incorporate this knowledge into the pseudo-labels provided by CLIP, thereby enhancing label quality. This explains why PLR delivers more substantial performance improvements in PDA.

\begin{figure}[!th]
\centering
\includegraphics[scale=0.32]{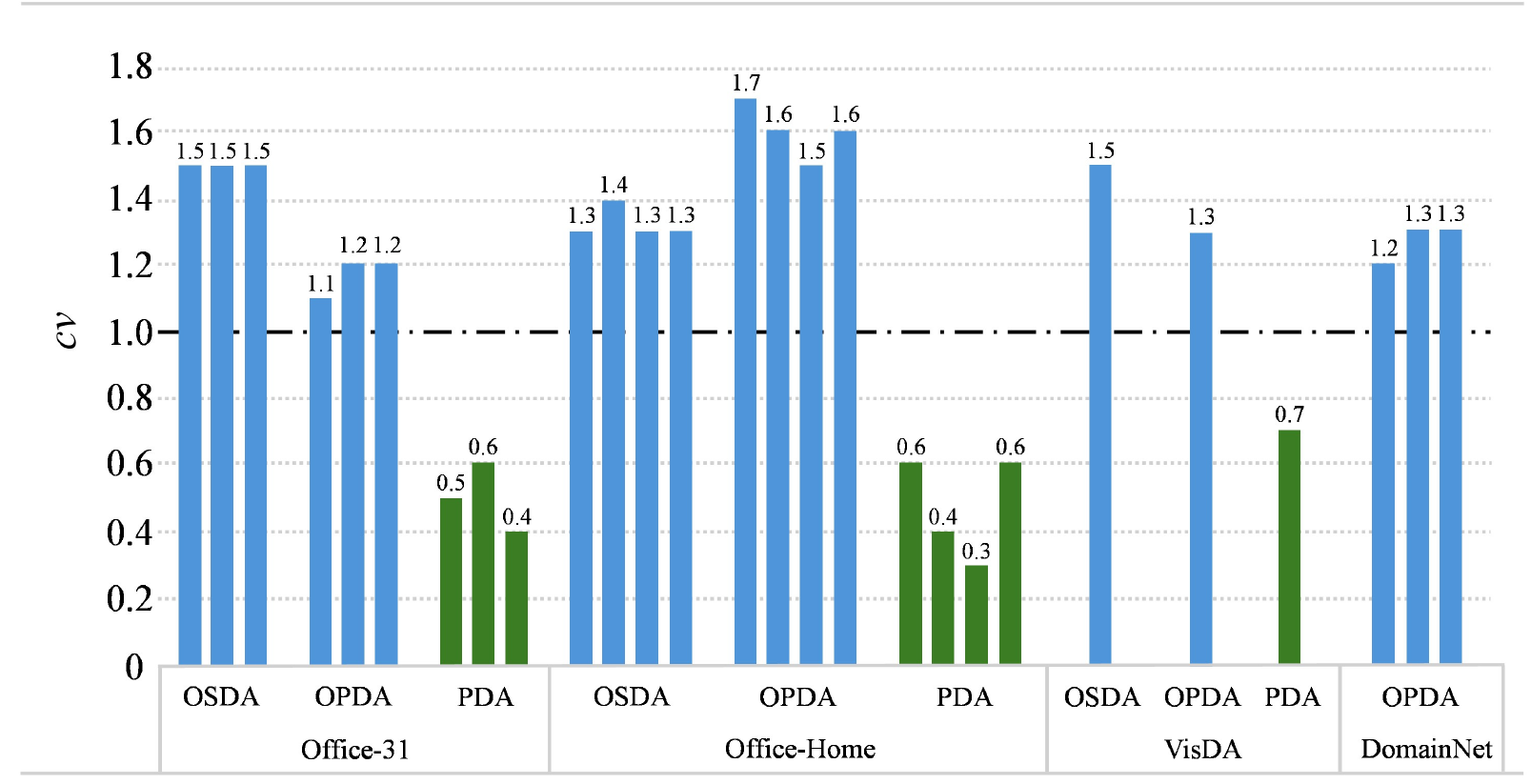}
\caption{Rationality of $cv$. Each bar represents a target domain, with digits showing $cv$ values under different label shifts. 
}
\label{fig5}
\vspace{-0.1in}
\end{figure}

\noindent\textbf{Rationality of Coefficient of Variation.} 
As shown in Fig.~\ref{fig5}, when the label shift corresponds to the PDA setting, the coefficient of variation ($cv$) across different target domains on the four benchmarks remains relatively low, with a maximum value of 0.7. In contrast, under the OPDA and OSDA settings which involve unknown categories, $cv$ is significantly higher, with a minimum of 1.1. Based on these observations, we set $\delta = 1.0$ as a threshold for $cv$ to distinguish between the presence or absence of unknown classes in the target domain. Specifically, if $cv > \delta$, this indicates the existence of unknown categories, implying that the label shift falls under the OPDA or OSDA setting. Otherwise, all target classes are assumed to be known from the source domain, suggesting a PDA setting.
{Notably, any choice of $\delta$ within the interval $[0.7, 1.1]$ leads to identical label-shift type determination across all benchmarks and target domains considered, indicating that the proposed criterion is not sensitive to the exact threshold value.}

\noindent\textbf{Sensitivity to the Percentage of Scores Used for $cv$ Computation.}
We further analyze the sensitivity of $cv$ to the percentage of selected scores used for its computation. Specifically, we vary the percentage from 10\% to 50\% and evaluate the resulting $cv$ values on Office-Home under three label shift settings. As reported in Table~\ref{tab:ScorePercentage}, when the target domain contains unknown classes (OSDA and OPDA), the $cv$ values consistently exceed 1.0 around the 30\% setting, whereas under PDA, where no unknown classes are present, the $cv$ values remain below 1. This clear separation validates both the choice of selecting the top and bottom 30\% scores for $cv$ computation and the use of a fixed threshold $\delta = 1.0$ for reliable label shift type determination.

\begin{figure}[!tp]
    \centering
    \includegraphics[scale=0.28]{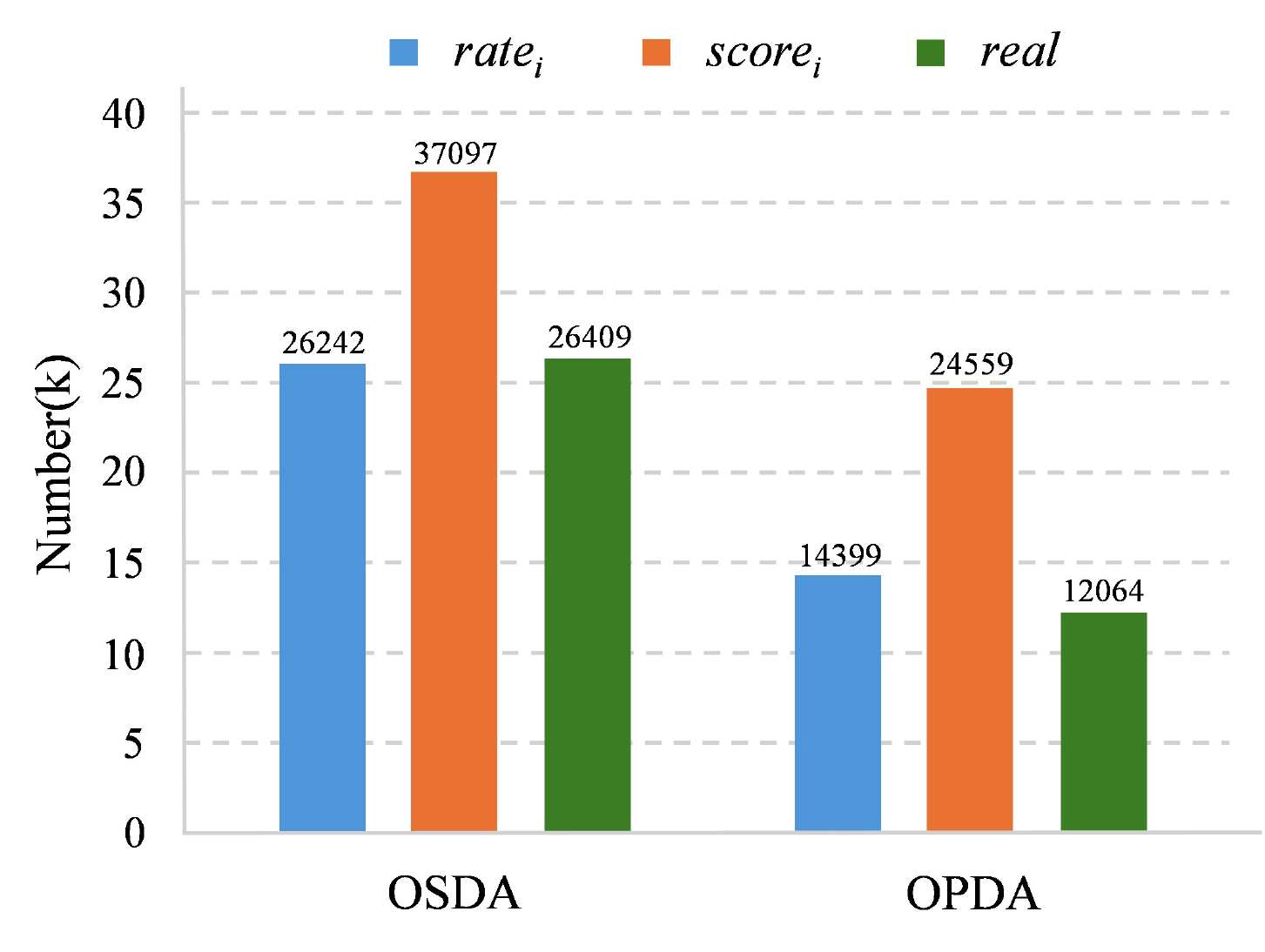}
    \setlength{\abovecaptionskip}{-0.01 cm}
    \caption{$score_i$ vs. $rate_i$. Comparison of GMM-based unknown sample separation using $score_i$ and $rate_i$ on VisDA. \textit{Real} denotes the actual number of unknown samples.}
    \label{fig:rateVSscore}
    \vspace{-0.1in}
\end{figure}

\noindent\textbf{Metric Comparison: $score_i$ versus $rate_i$.} In subsection~\ref{section:GMM}, we discuss the factors that cause the GMM to misclassify known samples as unknown. Fig.~\ref{fig:rateVSscore} provides empirical evidence supporting this claim. As illustrated in the figure, when GMM utilizes $score_i$ as the metric, the number of target samples identified as unknown is significantly greater than the actual number of target-private samples. In contrast, when $rate_i$ is used, the number of samples identified as unknown more closely aligns with the true count of target-private categories. This indicates that using $rate_i$ instead of $score_i$ improves the accuracy of unknown sample detection, thereby validating the rationale behind our choice of metric.

\begin{table}[!tp]\renewcommand\arraystretch{1.2}
    \centering
    \scalebox{0.9}{
    \begin{tabular}{l|c|c|c|c}
        \toprule
        \shortstack[l]{Benchmark\\(Domain)}  & \shortstack{Office\\(A)} & \shortstack{Office-Home\\(Cl)} & \shortstack{VisDA\\(R)} & \shortstack{DomainNet\\(S)}\\ 
        \midrule
        Time(ms)   & 22360  & 86170       & 291540 & 704510    \\ 
        Memory(MB) & 9634   & 9625        & 9524   & 13712     \\ 
        \bottomrule  
    \end{tabular}
    }
    \caption{Running efficiency of LFM under the OPDA setting. Time measures the total runtime, {including both one-time preprocessing components (LLM-based unknown-label generation, CLIP similarity computation, LSTD, and GMM-based unknown detection) and a representative training step consisting of one round of PLR, one training epoch, and one evaluation phase.} Memory denotes GPU memory usage.}
    \label{tab:efficiency}
\end{table}  

\begin{table}[!tp]\renewcommand\arraystretch{1.2}
    \centering
    \scalebox{0.9}{
\begin{tabular}{l|c|c|c|c}
    \toprule
    \shortstack[l]{Benchmark\\(Domain)}  
    & \shortstack{Office\\(A)} 
    & \shortstack{Office-Home\\(Cl)} 
    & \shortstack{VisDA\\(R)} 
    & \shortstack{DomainNet\\(S)}\\ 
    \midrule
    GLC \cite{qu2023upcycling} 
        & 543850  & 608660  & 704410  & 9919530 \\ 
    LFM (ours)                 
        & 3895    & 7960     & 58413   & 124147   \\

    {\footnotesize\quad -- LSTD} 
        & {\footnotesize 42} 
        & {\footnotesize 39} 
        & {\footnotesize 337} 
        & {\footnotesize 605} \\

    {\footnotesize\quad -- GMM}                  
        & {\footnotesize 433} 
        & {\footnotesize 1521} 
        & {\footnotesize 6116} 
        & {\footnotesize 7652} \\ 

    {\footnotesize\quad -- PLR}                  
        & {\footnotesize 3420} 
        & {\footnotesize 6400} 
        & {\footnotesize 51960} 
        & {\footnotesize 115890} \\  

    \bottomrule  
\end{tabular}
    }
    \label{tab_1}
    \caption{Time consumption comparison (ms) of one round of pseudo-label generation between LFM and the representative clustering-based SF-UniDA method GLC, measured on the target domain with the largest sample size for each benchmark. {Note that, the reported runtime focuses only on the pseudo-labeling components, excluding one-time preprocessing components, such as LLM-based unknown-label generation and CLIP similarity computation.}
    }
    \label{tab:runtime-glc}
    \vspace{-2em}
\end{table} 

\noindent\textbf{Running Efficiency.}
For a conservative evaluation, the running efficiency is measured on the domain with the largest number of samples in each benchmark. Table~\ref{tab:efficiency} summarizes the running efficiency of LFM, reporting the total time required for one-off processing (LLM-based unknown label generation, CLIP similarity computation, LSTD, and GMM-based unknown detection) together with a representative training step consisting of one round of PLR, one training epoch, and one evaluation, rather than the whole training process. GPU memory usage is also reported. As foundation models are not involved during the training of the target model, both time and memory consumption remain moderate in practice.

To provide a fair and focused efficiency comparison with clustering-based pseudo-labeling methods, we report the runtime of the pseudo-labeling strategy itself, rather than the end-to-end pipeline. Specifically, we exclude the time for unknown-class label generation via the LLM and similarity computation using CLIP, as these components are orthogonal to the pseudo-label operation and are not executed per training epoch. The LLM is queried only once for each label-shift type on a given benchmark to generate text labels for potential unknown categories, and its runtime is mainly affected by network conditions and LLM availability. Similarly, the CLIP-based similarity computation is performed only once before training the target model, which is independent of the pseudo-label assignment strategy. Including these components would therefore compare different paradigms rather than the efficiency of pseudo-labeling mechanisms. We emphasize that this comparison focuses on the efficiency of pseudo-labeling strategies rather than providing a full end-to-end runtime comparison of complete pipelines.

The reported runtime of LFM includes three core components, namely LSTD, GMM-based unknown class identification, and PLR. For the representative clustering-based method GLC \cite{qu2023upcycling}, we measure the time consumed by its clustering and pseudo-label assignment procedure. Consistent with the running efficiency evaluation, all measurements are performed on the domain with the largest sample size in each benchmark.

As shown in Table~\ref{tab:runtime-glc}, the time consumption of LFM for pseudo-label generation is reduced by one to two orders of magnitude compared to GLC, with PLR being the dominant cost while remaining lightweight. These results indicate that our pseudo-labeling strategy avoids the heavy computational overhead introduced by clustering. Moreover, the neighborhood soft-voting within the PLR component can be further accelerated using ANN libraries \cite{johnson2019billion} based implementations for improved scalability on large-scale datasets.

\noindent\textbf{{Complexity Discussion.}}
We analyze the computational complexity of each component in our framework and clarify their respective roles. We note that CLIP-based cross-modal similarity is computed using frozen, pre-trained encoders and is independent of the pseudo-labeling strategy itself. This step is performed in a shared, non-iterative manner and does not involve neighborhood construction. Therefore, it is orthogonal to the pseudo-labeling mechanism and is excluded from the method-specific complexity analysis.

The LSTD module operates on the domain-level statistic $cv$ and incurs a negligible overhead compared to the sample-wise similarity computation in the neighborhood soft-voting step. The GMM-based unknown class identification is performed on sample-specific $rate$, resulting in a complexity of $O(N_t)$. The PLR module introduces a per-epoch optimization cost that scales linearly with $N_t$.

For each target sample, the neighborhood soft-voting identifies its $K$ nearest neighbors based on cosine distance in feature space and averages their predicted probability vectors as defined in Eq.~\eqref{eq:neigborVote}. Computing pairwise cosine similarities requires $O(N_t^2 d)$ operations in the worst case, where $d$ denotes the feature dimension. The subsequent top-$K$ neighbor selection adds an $O(N_t K)$ cost. Since $K \ll N_t$ in practice, the dominant cost of this step is the similarity computation, which is performed once per PLR round (per epoch) and is comparable to prior neighborhood-based strategies\cite{liang2020we}. For large-scale target domains, this cost can be significantly reduced using ANN-based implementations such as FAISS \cite{johnson2019billion}, which provide efficient similarity approximation in high-dimensional spaces and ensure the scalability of our method.

Overall, the computational complexity of LFM is dominated by the KNN-based neighborhood soft-voting {under the current implementation}, while other components introduce only minor additional overhead, which is consistent with the runtime comparison in Table~\ref{tab:runtime-glc}.

\begin{figure}[!bp]
    \centering
    \includegraphics[scale=0.76]{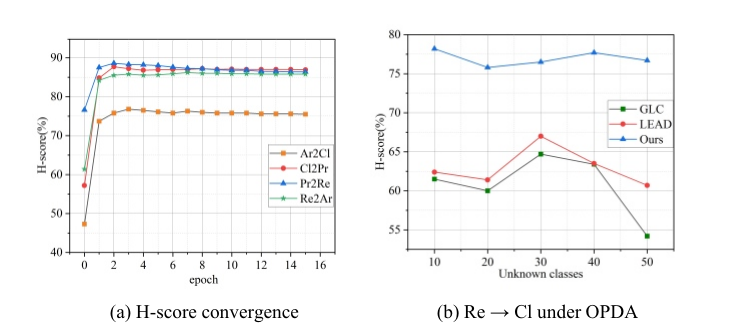}
    \setlength{\abovecaptionskip}{-0.001 cm}
    \caption{(a) H-score curves for 4 tasks on Office-Home under the OPDA setting. (b) Robustness comparison on the Rl→Cl task as the number of target-private categories increases.}
    \label{fig:sta_robus}
\end{figure}

\begin{figure*}[!tp]
        \centering
        \includegraphics[scale=0.68]{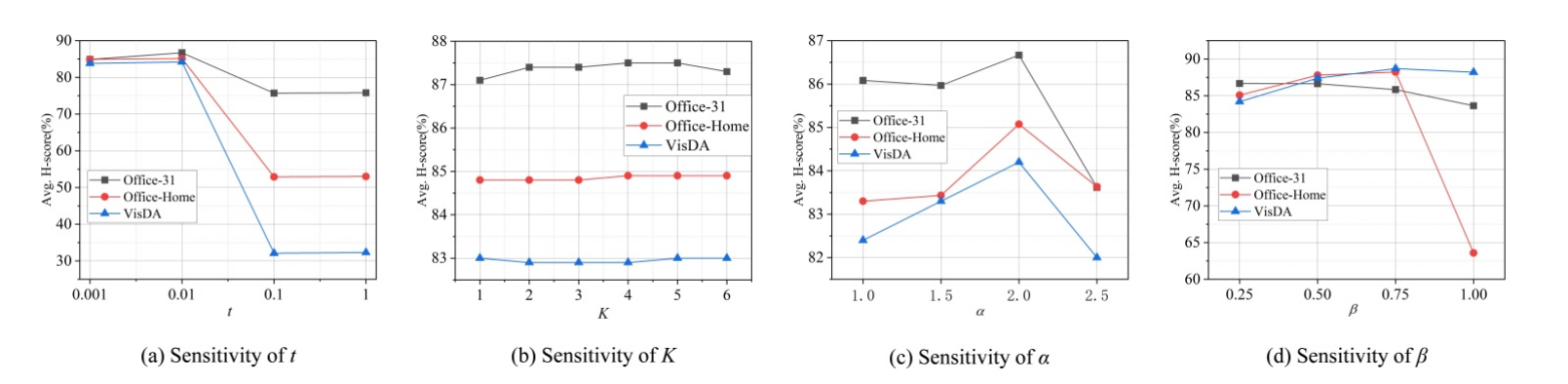}
        \setlength{\abovecaptionskip}{-0.5 cm}
        \caption{Hyperparameter sensitivity under the OPDA setting. (a) Sensitivity to scaling factor $t$. (b) $K$ has minimal impact. (c) Optimal $\alpha$ reflects target-private class ratio. (d) $\beta$ is benchmark-specific.}
        \label{fig:para_sens}
\end{figure*}

\begin{table*}[!tp]
\centering
\resizebox{\textwidth}{!}{
\begin{tabular}{l c >{\RaggedRight\arraybackslash}m{5.2cm} c >{\RaggedRight\arraybackslash}m{6.5cm}}
\hline
\textbf{Benchmark} & \textbf{\#Known} & \textbf{Known-class text labels} & \textbf{\#Unknown} & \textbf{Generated unknown-class text labels} \\
\hline
Office-31 & 20 &
back\_pack, bike, calculator, headphones, keyboard, laptop\_computer, monitor, mouse, mug, projector, 
bike\_helmet, bookcase, bottle, desk\_chair, desk\_lamp, desktop\_computer, file\_cabinet, letter\_tray, mobile\_phone, paper\_notebook
& 40 &
 carabiner, motorcycle, abacus, earmuffs, skateboard, piano\_keys, tablet\_device, television\_screen, joystick, teacup, cinema\_screen,
sports\_helmet, thermos\_flask, armchair, standing\_lamp, gaming\_console, filing\_rack, organizer\_box, diary\_book, suitcase, whiteboard, office\_sofa, wall\_mirror, video\_camera, gaming\_controller, coffee\_pot, projector\_screen,
slide\_rule, radio\_receiver, accordion, e\_reader, riding\_helmet, storage\_rack,
laboratory\_flask, , mini\_computer, drawer\_unit, document\_holder, flip\_phone, travel\_journal,  desk\_organizer 
\\
\hline
Office-Home & 15 &
Alarm\_Clock, Backpack, Batteries, Bed, Bike, Bottle, Bucket, Calculator, Calendar, Candles, Chair, Clipboards, Computer, Couch, Curtains
& 30 &
Desk, Wallet, Writing\_Notebook, Floor\_Lamp, Pen, Stove, Blender, Wall\_Mirror, Pillow, Scooter,
Travel\_Mug, Bathtub, Weighing\_Scale, Wristwatch, Window\_Blinds, Television, Bench, File\_Folder, Tablet\_PC, Armchair, Dining\_Table, Electric\_Kettle, Picture\_Frame, Decorative\_Vase, Toaster\_Oven, Bookshelf\_Unit, Hammer\_Tool, Screwdriver\_Tool, Adjustable\_Wrench, Tool\_Box 
\\
\hline
VisDA & 9 &
aeroplane, bicycle, bus, car, horse, knife, motorcycle, person, plant
& 18 &
helicopter, skateboard, train, truck, camel, scissors, scooter, statue, tree\_trunk, ship, rollerblade, cable\_car, zebra, sword, mannequin, shrub, sailing\_boat, traffic\_cone 
\\
\hline
DomainNet & 200 &
e.g., airplane, bicycle, dog, chair, cup, guitar, helicopter, pizza, rabbit, television
& 400 &
e.g., volcano, crystal\_ball, skyscraper, iceberg, tornado, windmill, coral\_reef, waffle\_iron, lava\_lamp, origami\_crane  
\\
\hline
\end{tabular}
}
\caption{Known-class text labels and representative examples of LLM-generated unknown-class text labels under the OPDA setting. For Office-31, Office-Home, and VisDA, all known and unknown class labels are reported, whereas for the large-scale benchmark DomainNet, only representative examples of both known and unknown class labels are shown for clarity.}
\label{tab:unknown-label-examples}
\end{table*}

\noindent\textbf{Stability and Robustness Analysis.}
Fig.~\ref{fig:sta_robus} (a) shows that the training of LFM converges rapidly and remains stable thereafter. Fig.~\ref{fig:sta_robus} (b) illustrates the effect of increasing the number of target-private categories on the performance of different methods. As the number of unknown categories grows, the adaptation task becomes increasingly difficult. Using the Rl→Cl adaptation on Office-Home under the OPDA setting as the evaluation task, LFM consistently outperforms state-of-the-art methods GLC and LEAD, maintaining stable performance and demonstrating strong robustness to the presence of an increasing number of unknown classes.

\noindent\textbf{Parameter Sensitivity.}
Fig.~\ref{fig:para_sens} presents a comprehensive sensitivity analysis of key hyperparameters, validating the appropriateness of the chosen settings.
When evaluating the sensitivity of a specific hyperparameter, all remaining hyperparameters are fixed to their default values ($t = 0.01$, $K = 5$, $\alpha = 2$, and $\beta = 0.25$) to ensure a controlled and reproducible evaluation. Subfigure (a) supports our selection of $t = 0.01$ across all benchmarks. Subfigure (b) shows that $K$, the number of neighbors used in the neighborhood soft voting strategy, has minimal impact on performance, justifying the fixed setting of $K = 5$ in all experiments. In subfigure (c), the average H-score peaks when the number of target-private classes is twice that of the common categories, supporting the choice of $\alpha = 2$. As shown in subfigure (d), the optimal value of $\beta$ is 0.25 for Office-31, whereas for Office-Home and VisDA, the best results are obtained with $\beta = 0.75$, indicating that $\beta$ is a benchmark-specific hyperparameter. Nevertheless, $\beta$ is fixed to the default value when analyzing the sensitivity of the other hyperparameters, following the control-variable principle. We note that possible interaction effects among hyperparameters were not exhaustively explored in this work.

\noindent
\textbf{Examples of Generated Unknown-class Text Labels.}
To further clarify the behavior of the LLM-based unknown category generation, we provide representative examples of the envisioned unknown-class text labels for each benchmark.
It is worth noting that the generated unknown labels are conditioned on the set of known class labels provided to the LLM, and their number is set to twice the number of known classes. Therefore, the exact unknown label sets vary across benchmarks and label shift types. We focus on OPDA because it involves both shared classes and source-private classes, resulting in a non-trivial unknown-category space in the target domain. The known-class labels and the corresponding representative examples of LLM-generated unknown-class labels under the OPDA setting are reported in Table~\ref{tab:unknown-label-examples}.

\begin{table}[!tp]\renewcommand\arraystretch{1.2} 
\centering
\scalebox{0.75}{
    \begin{tabular}{l|ccccccc|c}
        \toprule
        \multirow{2}[1]{*}{Methods} & \multicolumn{7}{c|}{Office-31} & \multicolumn{1}{c}{VisDA}\\
        \cline{2-9} &A2D &A2W &D2A &D2W &W2A &W2D &\textbf{Avg.} & \textbf{S2R}\\
        \hline
        CLIP Zero-Shot$^{\dagger}$  
                   &60.1  &57.3  &57.0  &57.3 &57.0  &60.1  &58.1 &61.5 \\
        LLM+CLIP   &80.8 &75.1  &80.9  &75.1 &80.9  &80.8  &78.9 &75.8 \\
        LFM (dissimilar) &82.1 &80.4 &84.4  &82.3 &84.8 &89.1 &83.9 &78.0 \\
        LFM (irrelevant) &87.1 &81.5 &85.5  &83.0 &85.2 &92.0 &85.7 &78.9 \\
        \hline
        \rowcolor{Gray}LFM (Ours) &\textbf{89.4} 
                   &\textbf{81.9} &\textbf{88.0}  &\textbf{83.9} &\textbf{88.1} &\textbf{94.0} &\textbf{87.6} &\textbf{86.2} \\
        \hline
    \end{tabular}
}
\caption{{H-score comparison when the proposed LFM using visually ``dissimilar'' or ``irrelevant'' prompts for the LLM. The experiments are conducted under the OPDA setting. 
} 
}
\vspace{-0.2in}
\label{tab:nonsimilarPromps} 
\end{table}

\noindent\textbf{Effectiveness of the Envisioned Unknown Class Labels.}
Target domain data in the real world is often highly diverse and unpredictable, so it is unrealistic to expect the generated labels to perfectly match the ground truth (GT) target-private classes. The purpose of generating potential unknown-class labels is not to accurately match the GT classes, but to assist in the detection of unknown target classes. 
Our approach leverages a visual similarity rule, which ensures that even if the generated unknown labels do not correspond exactly to the GT unknown classes, they still help identify unknown categories based on visual similarity. We provide a specific example under OPDA on Office-31, where the LLM generates the label `video$\_$camera' as an unknown class. Although the generated label does not perfectly match the target private class `projector', the `video$\_$camera' exhibits a high visual similarity to `projector' (both have a rectangular box shape and a lens). As a result, `projector' could be classified as `video$\_$camera', an unknown class. Otherwise, `projector' would be misclassified as one of the known classes, leading to degraded performance.

To investigate the effectiveness of this visual similarity rule, we conduct additional experiments with two modified LLM prompts: one with a `visually irrelevant' constraint and another with a `visually dissimilar' constraint. The `visually irrelevant' prompt generates unknown-class labels without considering visual similarity, while the `visually dissimilar' prompt generates labels under a constraint that is visually dissimilar to the source-known classes. As shown in Table~\ref{tab:nonsimilarPromps}, the classification performance without the visual similarity constraint is significantly worse, underscoring the importance of the visual similarity-based unknown-class generation strategy. However, even with the degraded results, our method still outperforms CLIP Zero-Shot$^{\dagger}$ and LLM+CLIP, emphasizing the strength of our complete framework, which includes components such as LSTD, GMM-based unknown-class separation, and PLR.

\subsection{Visualization}
\begin{figure}[!tp]
    \centering
    \includegraphics[scale=0.33]{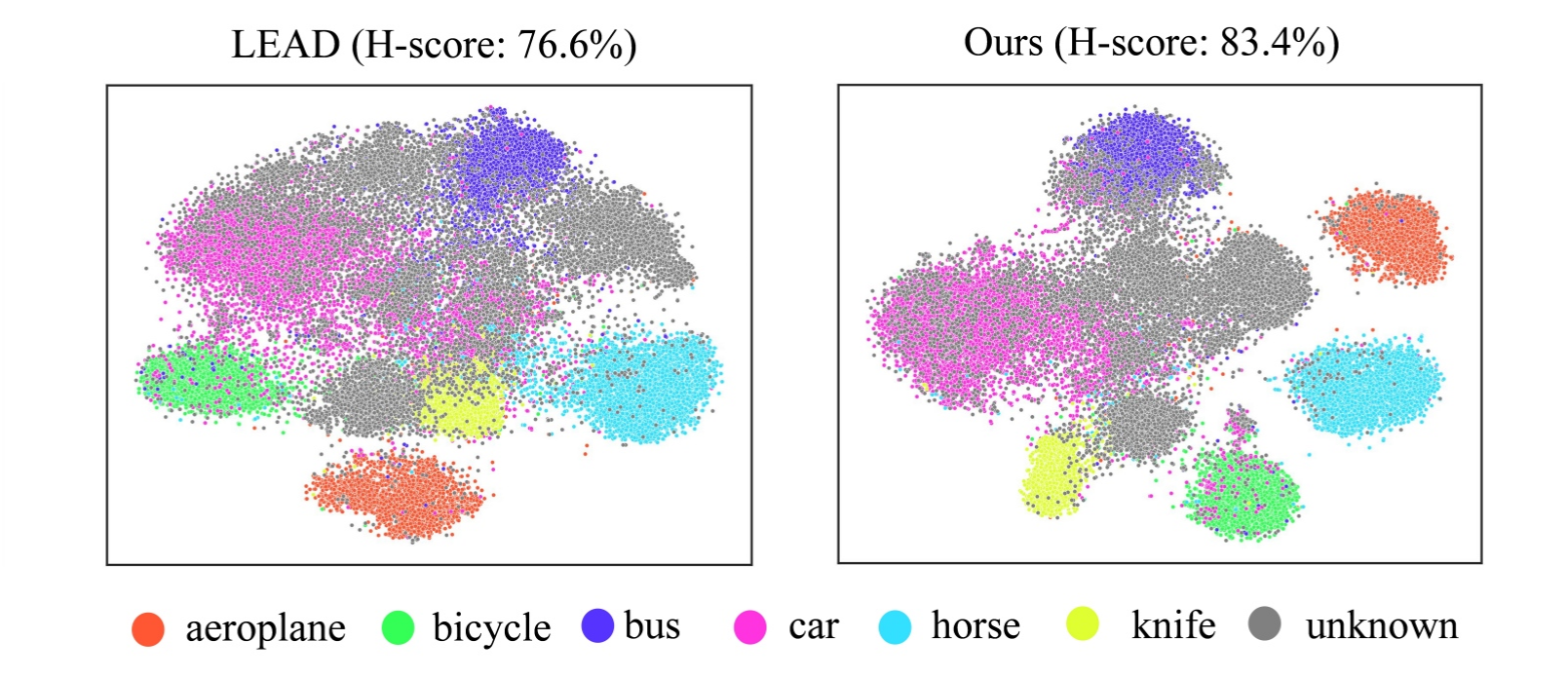}
    \caption{t-SNE visualization for the OPDA task S→R on VisDA.
    }
    \label{fig:tsne}
    \vspace{-0.1in}
\end{figure}

\begin{figure}[!t]
    \centering
    \includegraphics[scale=0.33]{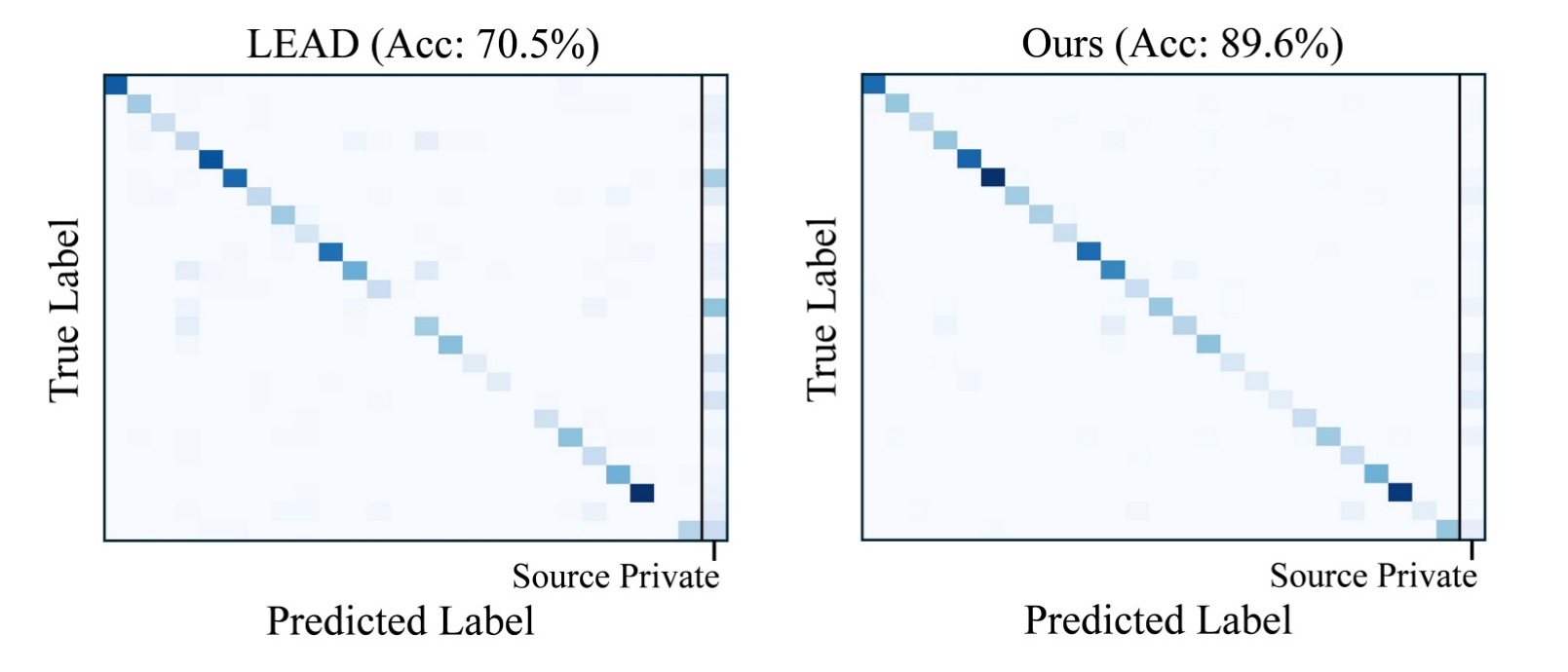}
    \setlength{\abovecaptionskip}{-0.5 cm}
    \caption{Confusion matrices for the PDA task Cl→Ar.}
    \label{fig:cm}
    \vspace{-0.1in}
\end{figure}

\begin{figure}[!t]
    \centering
    \includegraphics[scale=0.32]{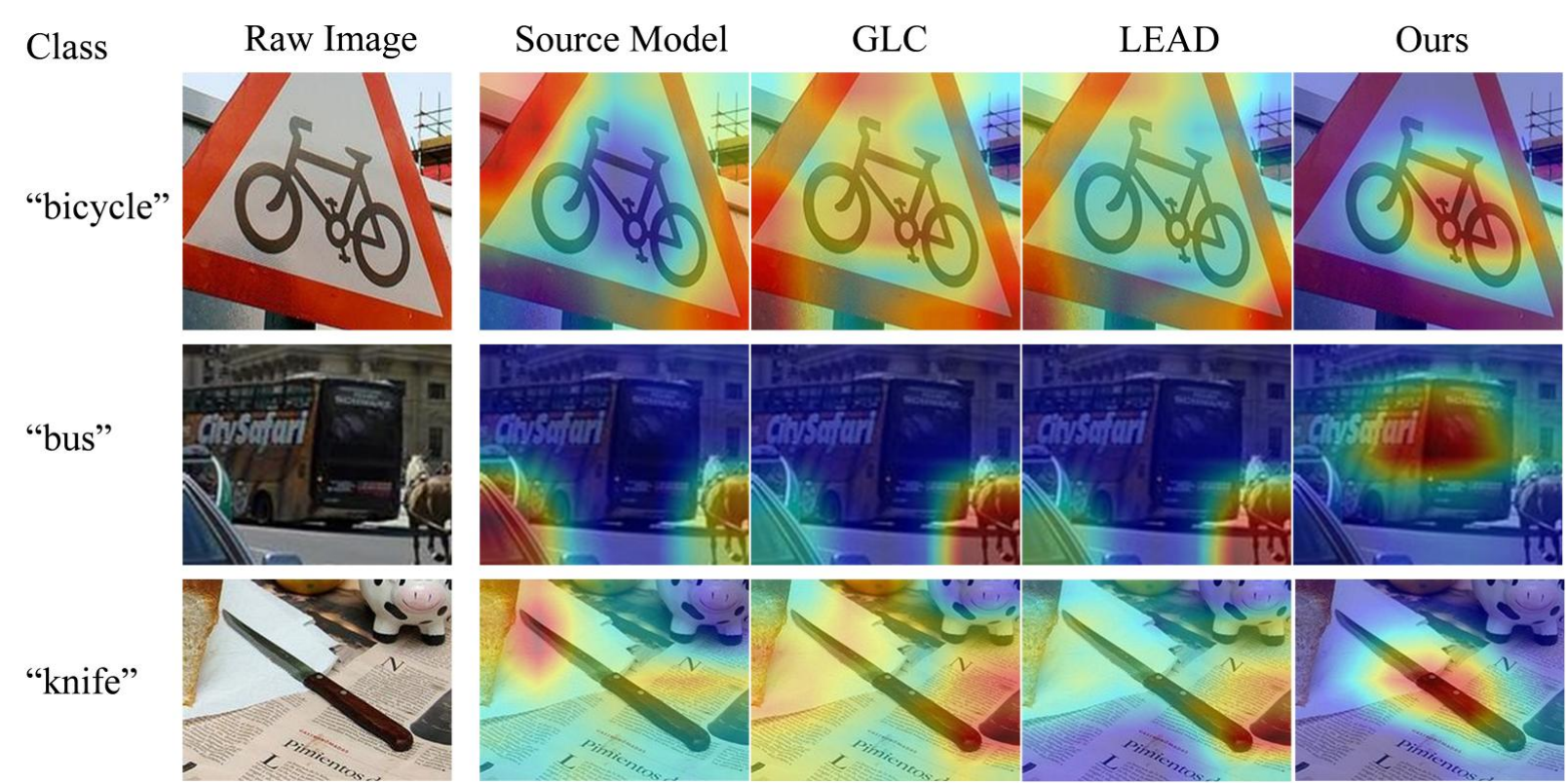}
    \caption{Visualization of regions vital for classification. 
    }
    \label{fig:cam}
    \vspace{-0.2in}
\end{figure}

\noindent\textbf{Feature Space Visualization.}
For the S→R task under the OPDA setting on VisDA, t-SNE \cite{van2008visualizing} is used to visualize the target domain feature space extracted by LEAD and our method. As shown in Fig.~\ref{fig:tsne}, our method yields more discriminative representations, demonstrating stronger classification of known categories and better separation of unknown samples. 

\noindent\textbf{Confusion Matrix.}
For the PDA task Cl→Ar on Office-Home, Fig.~\ref{fig:cm} presents the confusion matrices based on the predictions made by LEAD and our method. Integrating knowledge from both CLIP and the target model, LFM reduces misclassifications, showing its ability to adapt under the PDA scenario.

\noindent\textbf{Vital Regions Visualization.}
Fig.~\ref{fig:cam} uses Grad-CAM \cite{selvaraju2017grad} to visualize the image regions attended to by different models during classification. Red regions indicate high importance, while purple denotes low relevance. The first column shows category names, followed by raw images from VisDA, and attention maps from four methods. Compared to the Source Model, LFM consistently focuses on discriminative regions, whereas GLC and LEAD often miss key areas. 

\section{Conclusion}
In this paper, we propose a unified and practical SF-UniDA framework named LFM. Our approach leverages foundation models, such as LLMs and VLMs, to effectively adapt the pre-trained source model to the target domain. We introduce two sample-level metrics to determine the label shift types and separate the unknown samples. Moreover, we refine the pseudo-labels generated by CLIP using the knowledge of the target model initialized from the pre-trained source model, thereby integrating complementary insights from both foundation models and the source domain. Extensive experiments show that LFM achieves competitive and consistently strong performance across diverse SF-UniDA scenarios. Our findings suggest that foundation models can serve as effective semantic priors for SF-UniDA when coupled with explicit label-shift inference and target-adaptive refinement, and may encourage further investigation into principled FM-assisted solutions for realistic source-free adaptation settings.

\normalem
\bibliographystyle{IEEEtran}
\bibliography{ref}

@ARTICLE{DPL2025TMM,
  author={Liu, Pan and Li, Jing and Zhao, Meng and Xue, Wanli and Hu, Qinghua and Chen, Shengyong},
  journal={IEEE Transactions on Multimedia}, 
  title={Domain-Division Based Progressive Learning for Source-Free Domain Adaptation}, 
  year={2025},
  volume={27},
  number={},
  pages={7081-7092},
  doi={10.1109/TMM.2025.3590903}}

@inproceedings{he2016deep,
  title={Deep residual learning for image recognition},
  author={He, Kaiming and Zhang, Xiangyu and Ren, Shaoqing and Sun, Jian},
  booktitle={Proceedings of the IEEE conference on computer vision and pattern recognition},
  pages={770--778},
  year={2016}
}

@inproceedings{deng2025multi,
  title={Multi-granularity class prototype topology distillation for class-incremental source-free unsupervised domain adaptation},
  author={Deng, Peihua and Zhang, Jiehua and Sheng, Xichun and Yan, Chenggang and Sun, Yaoqi and Fu, Ying and Li, Liang},
  booktitle={Proceedings of the Computer Vision and Pattern Recognition Conference},
  pages={30566--30576},
  year={2025}
}

@InProceedings{Antonio2024CDS,
author="Tejero-de-Pablos, Antonio
and Togashi, Riku
and Otani, Mayu
and Satoh, Shin'ichi",
editor="Leonardis, Ale{\v{s}}
and Ricci, Elisa
and Roth, Stefan
and Russakovsky, Olga
and Sattler, Torsten
and Varol, G{\"u}l",
title="Robust Nearest Neighbors for Source-Free Domain Adaptation Under Class Distribution Shift",
booktitle="Computer Vision -- ECCV 2024",
year="2025",
publisher="Springer Nature Switzerland",
address="Cham",
pages="1--17",
isbn="978-3-031-73220-1"
}

@InProceedings{eccv2024DCPL,
author="Diamant, Idit
and Rosenfeld, Amir
and Achituve, Idan
and Goldberger, Jacob
and Netzer, Arnon",
editor="Leonardis, Ale{\v{s}}
and Ricci, Elisa
and Roth, Stefan
and Russakovsky, Olga
and Sattler, Torsten
and Varol, G{\"u}l",
title="De-confusing Pseudo-labels in Source-Free Domain Adaptation",
booktitle="Computer Vision -- ECCV 2024",
year="2025",
publisher="Springer Nature Switzerland",
address="Cham",
pages="108--125",
isbn="978-3-031-72986-7"
}

@inproceedings{xing2024hierarchical,
  title={Hierarchical Unsupervised Relation Distillation for Source-Free Domain Adaptation},
  author={Xing, Bowei and Ying, Xianghua and Wang, Ruibin and Guo, Ruohao and Shi, Ji and Yue, Wenzhen},
  booktitle={European Conference on Computer Vision},
  pages={393--409},
  year={2024},
  organization={Springer}
}

@InProceedings{Hao2024ECCV,
author="Hao, Yan
and Forest, Florent
and Fink, Olga",
editor="Leonardis, Ale{\v{s}}
and Ricci, Elisa
and Roth, Stefan
and Russakovsky, Olga
and Sattler, Torsten
and Varol, G{\"u}l",
title="Simplifying Source-Free Domain Adaptation for Object Detection: Effective Self-training Strategies and Performance Insights",
booktitle="Computer Vision -- ECCV 2024",
year="2025",
publisher="Springer Nature Switzerland",
address="Cham",
pages="196--213",
isbn="978-3-031-72949-2"
}

@article{johnson2019billion,
  title={Billion-scale similarity search with {GPUs}},
  author={Johnson, Jeff and Douze, Matthijs and J{\'e}gou, Herv{\'e}},
  journal={IEEE Transactions on Big Data},
  volume={7},
  number={3},
  pages={535--547},
  year={2019},
  publisher={IEEE}
}

@inproceedings{tang2024source,
  title={Source-free domain adaptation with frozen multimodal foundation model},
  author={Tang, Song and Su, Wenxin and Ye, Mao and Zhu, Xiatian},
  booktitle={Proceedings of the IEEE/CVF Conference on Computer Vision and Pattern Recognition},
  pages={23711--23720},
  year={2024}
}

@inproceedings{Monga_2024_BMVC,
author    = {Munish Monga and Sachin Kumar Giroh and Ankit Jha and Mainak Singha and Biplab Banerjee and Jocelyn Chanussot},
title     = {COSMo: CLIP Talks on Open-Set Multi-Target Domain Adaptation},
booktitle = {35th British Machine Vision Conference 2024, {BMVC} 2024, Glasgow, UK, November 25-28, 2024},
publisher = {BMVA},
year      = {2024},
url       = {https://papers.bmvc2024.org/0031.pdf}
}

@INPROCEEDINGS{PromptDIV2024ICIP,
  author={Zeng, Shihao and Liu, Xinghong and Zhou, Yi},
  booktitle={2024 IEEE International Conference on Image Processing (ICIP)}, 
  title={Decoupling Domain Invariance and Variance With Tailored Prompts for Open-Set Domain Adaptation}, 
  year={2024},
  volume={},
  number={},
  pages={645-651},
  doi={10.1109/ICIP51287.2024.10647719}}

@article{yu2025open,
  title={Open-set domain adaptation with visual-language foundation models},
  author={Yu, Qing and Irie, Go and Aizawa, Kiyoharu},
  journal={Computer Vision and Image Understanding},
  volume={250},
  pages={104230},
  year={2025},
  publisher={Elsevier}
}

@article{deng2023universal,
  title={Universal domain adaptation from foundation models: A baseline study},
  author={Deng, Bin and Jia, Kui},
  journal={arXiv preprint arXiv:2305.11092},
  year={2023}
}

@inproceedings{zhu2023UniAM,
  title={Universal domain adaptation via compressive attention matching},
  author={Zhu, Didi and Li, Yinchuan and Yuan, Junkun and Li, Zexi and Kuang, Kun and Wu, Chao},
  booktitle={Proceedings of the IEEE/CVF International Conference on Computer Vision},
  pages={6974--6985},
  year={2023}
}

@article{zhang2025source,
  title={Source-free domain adaptation guided by vision and vision-language pre-training},
  author={Zhang, Wenyu and Shen, Li and Foo, Chuan-Sheng},
  journal={International Journal of Computer Vision},
  volume={133},
  number={2},
  pages={844--866},
  year={2025},
  publisher={Springer}
}

@article{wang2022cross,
  author={Wang, Rui and Wu, Zuxuan and Weng, Zejia and Chen, Jingjing and Qi, Guo-Jun and Jiang, Yu-Gang},
  journal={IEEE Transactions on Multimedia}, 
  title={Cross-Domain Contrastive Learning for Unsupervised Domain Adaptation}, 
  year={2023},
  volume={25},
  number={},
  pages={1665-1673}
}

@ARTICLE{Tian2024,
  author={Tian, Yuntong and Li, Jiaxi and Fu, Huazhu and Zhu, Lei and Yu, Lequan and Wan, Liang},
  journal={IEEE Transactions on Multimedia}, 
  title={Self-Mining the Confident Prototypes for Source-Free Unsupervised Domain Adaptation in Image Segmentation}, 
  year={2024},
  volume={26},
  number={},
  pages={7709-7720},
  doi={10.1109/TMM.2024.3370678}
  }

@inproceedings{wen2024cross,
  title={Cross-domain Open-world Discovery},
  author={Wen, Shuo and Brbic, Maria},
  booktitle={International Conference on Machine Learning},
  pages={52744--52761},
  year={2024},
  organization={PMLR}
}

@ARTICLE{DomainPromptTuning,
  author={Jin, Xin and Lan, Cuiling and Zeng, Wenjun and Chen, Zhibo},
  journal={IEEE Transactions on Multimedia}, 
  title={Domain Prompt Tuning via Meta Relabeling for Unsupervised Adversarial Adaptation}, 
  year={2024},
  volume={26},
  number={},
  pages={8333-8347}
  }

@INPROCEEDINGS{jing2023ICME,
  author={Li, Jing and Yang, Liu and Wang, Qilong and Hu, Qinghua},
  booktitle={2023 IEEE International Conference on Multimedia and Expo (ICME)}, 
  title={Coarse Helps Fine: A Multi-Granularity Discriminative Adversarial Network for Fine-Grained Open-Set Domain Adaptation}, 
  year={2023},
  volume={},
  number={},
  pages={2675-2680},
  doi={10.1109/ICME55011.2023.00455}}

@ARTICLE{jing2023WDAN,
  author={Li, Jing and Yang, Liu and Wang, Qilong and Hu, Qinghua},
  journal={IEEE Transactions on Circuits and Systems for Video Technology}, 
  title={WDAN: A Weighted Discriminative Adversarial Network With Dual Classifiers for Fine-Grained Open-Set Domain Adaptation}, 
  year={2023},
  volume={33},
  number={9},
  pages={5133-5147},
  doi={10.1109/TCSVT.2023.3249200}}

@ARTICLE{jingNENO,
  author={Li, Jing and Yang, Liu and Hu, Qinghua},
  journal={IEEE Transactions on Circuits and Systems for Video Technology}, 
  title={Enhancing Multi-Source Open-Set Domain Adaptation Through Nearest Neighbor Classification With Self-Supervised Vision Transformer}, 
  year={2024},
  volume={34},
  number={4},
  pages={2648-2662},
  doi={10.1109/TCSVT.2023.3307789}}

@article{bommasani2021opportunities,
  title={On the opportunities and risks of foundation models},
  author={Bommasani, Rishi and Hudson, Drew A and Adeli, Ehsan and Altman, Russ and Arora, Simran and von Arx, Sydney and Bernstein, Michael S and Bohg, Jeannette and Bosselut, Antoine and Brunskill, Emma and others},
  journal={arXiv preprint arXiv:2108.07258},
  year={2021}
}

@article{li2018domain,
  title={Domain invariant and class discriminative feature learning for visual domain adaptation},
  author={Li, Shuang and Song, Shiji and Huang, Gao and Ding, Zhengming and Wu, Cheng},
  journal={IEEE transactions on image processing},
  volume={27},
  number={9},
  pages={4260--4273},
  year={2018},
  publisher={IEEE}
}

@inproceedings{kundu2020universal,
  title={Universal source-free domain adaptation},
  author={Kundu, Jogendra Nath and Venkat, Naveen and Babu, R Venkatesh and others},
  booktitle={Proceedings of the IEEE/CVF conference on computer vision and pattern recognition},
  pages={4544--4553},
  year={2020}
}

@article{voigt2017eu,
  title={The eu general data protection regulation (gdpr)},
  author={Voigt, Paul and Von dem Bussche, Axel},
  journal={A Practical Guide, 1st Ed., Cham: Springer International Publishing},
  volume={10},
  number={3152676},
  pages={10--5555},
  year={2017},
  publisher={Springer}
}

@article{coefficientofvariation,
title = {Monitoring the coefficient of variation: A literature review},
journal = {Computers \& Industrial Engineering},
volume = {161},
pages = {107600},
year = {2021},
issn = {0360-8352},
doi = {https://doi.org/10.1016/j.cie.2021.107600},
url = {https://www.sciencedirect.com/science/article/pii/S0360835221005040},
author = {Zahra Jalilibal and Amirhossein Amiri and Philippe Castagliola and Michael B.C. Khoo}
}

@ARTICLE{ZhangDecadeSurvey2022,
  author={Zhang, Lei and Gao, Xinbo},
  journal={IEEE Transactions on Neural Networks and Learning Systems}, 
  title={Transfer Adaptation Learning: A Decade Survey}, 
  year={2024},
  volume={35},
  number={1},
  pages={23-44},
  doi={10.1109/TNNLS.2022.3183326}}

@InProceedings{pmlr-v235-cao24d,
  title = {Envisioning Outlier Exposure by Large Language Models for Out-of-Distribution Detection},
  author = {Cao, Chentao and Zhong, Zhun and Zhou, Zhanke and Liu, Yang and Liu, Tongliang and Han, Bo},
  booktitle =    {Proceedings of the 41st International Conference on Machine Learning},
  pages =    {5629--5659},
  year =   {2024},
  volume =   {235},
  month =    {21--27 Jul},
  publisher =    {PMLR},
}

@inproceedings{vaswani2017attention,
 author = {Vaswani, Ashish and Shazeer, Noam and Parmar, Niki and Uszkoreit, Jakob and Jones, Llion and Gomez, Aidan N and Kaiser, \L ukasz and Polosukhin, Illia},
 booktitle = {Advances in Neural Information Processing Systems},
 editor = {I. Guyon and U. Von Luxburg and S. Bengio and H. Wallach and R. Fergus and S. Vishwanathan and R. Garnett},
 pages = {1-11},
 publisher = {Curran Associates, Inc.},
 title = {Attention is All you Need},
 volume = {30},
 year = {2017}
}

@inproceedings{xiao2024adversarial,
  title={Adversarial Experts Model for Black-box Domain Adaptation},
  author={Xiao, Siying and Ye, Mao and He, Qichen and Li, Shuaifeng and Tang, Song and Zhu, Xiatian},
  booktitle={Proceedings of the 32nd ACM International Conference on Multimedia},
  pages={8982--8991},
  year={2024}
}

@inproceedings{litrico2023guiding,
  title={Guiding pseudo-labels with uncertainty estimation for source-free unsupervised domain adaptation},
  author={Litrico, Mattia and Del Bue, Alessio and Morerio, Pietro},
  booktitle={Proceedings of the IEEE/CVF Conference on Computer Vision and Pattern Recognition},
  pages={7640--7650},
  year={2023}
}

@inproceedings{qu2023upcycling,
  title={Upcycling models under domain and category shift},
  author={Qu, Sanqing and Zou, Tianpei and R{\"o}hrbein, Florian and Lu, Cewu and Chen, Guang and Tao, Dacheng and Jiang, Changjun},
  booktitle={Proceedings of the IEEE/CVF Conference on Computer Vision and Pattern Recognition},
  pages={20019--20028},
  year={2023}
}

@inproceedings{qu2024lead,
  title={Lead: Learning decomposition for source-free universal domain adaptation},
  author={Qu, Sanqing and Zou, Tianpei and He, Lianghua and R{\"o}hrbein, Florian and Knoll, Alois and Chen, Guang and Jiang, Changjun},
  booktitle={Proceedings of the IEEE/CVF Conference on Computer Vision and Pattern Recognition},
  pages={23334--23343},
  year={2024}
}

@inproceedings{long2015learning,
  author = {Long, Mingsheng and Cao, Yue and Wang, Jianmin and Jordan, Michael I.},
  title = {Learning transferable features with deep adaptation networks},
  year = {2015},
  publisher = {JMLR.org},
  booktitle = {Proceedings of the 32nd International Conference on International Conference on Machine Learning - Volume 37},
  pages = {97–105},
  numpages = {9},
  location = {Lille, France},
  series = {ICML'15}
}

@inproceedings{long2017deep,
  title={Deep transfer learning with joint adaptation networks},
  author={Long, Mingsheng and Zhu, Han and Wang, Jianmin and Jordan, Michael I},
  booktitle={International conference on machine learning},
  pages={2208--2217},
  year={2017},
  organization={PMLR}
}

@article{ganin2016domain,
  title={Domain-adversarial training of neural networks},
  author={Ganin, Yaroslav and Ustinova, Evgeniya and Ajakan, Hana and Germain, Pascal and Larochelle, Hugo and Laviolette, Fran{\c{c}}ois and March, Mario and Lempitsky, Victor},
  journal={Journal of machine learning research},
  volume={17},
  number={59},
  pages={1--35},
  year={2016}
}

@inproceedings{radford2021learning,
  title={Learning transferable visual models from natural language supervision},
  author={Radford, Alec and Kim, Jong Wook and Hallacy, Chris and Ramesh, Aditya and Goh, Gabriel and Agarwal, Sandhini and Sastry, Girish and Askell, Amanda and Mishkin, Pamela and Clark, Jack and others},
  booktitle={International conference on machine learning},
  pages={8748--8763},
  year={2021},
  organization={PMLR}
}

@article{achiam2023gpt,
  title={Gpt-4 technical report},
  author={Achiam, Josh and Adler, Steven and Agarwal, Sandhini and Ahmad, Lama and Akkaya, Ilge and Aleman, Florencia Leoni and Almeida, Diogo and Altenschmidt, Janko and Altman, Sam and Anadkat, Shyamal and others},
  journal={arXiv preprint arXiv:2303.08774},
  year={2023}
}

@article{liu2022psdc,
  title={PSDC: A prototype-based shared-dummy classifier model for open-set domain adaptation},
  author={Liu, Zhengfa and Chen, Guang and Li, Zhijun and Kang, Yu and Qu, Sanqing and Jiang, Changjun},
  journal={IEEE Transactions on Cybernetics},
  volume={53},
  number={11},
  pages={7353--7366},
  year={2022},
  publisher={IEEE}
}

@inproceedings{cao2019learning,
  title={Learning to transfer examples for partial domain adaptation},
  author={Cao, Zhangjie and You, Kaichao and Long, Mingsheng and Wang, Jianmin and Yang, Qiang},
  booktitle={Proceedings of the IEEE/CVF conference on computer vision and pattern recognition},
  pages={2985--2994},
  year={2019}
}

@inproceedings{saito2021ovanet,
  title={Ovanet: One-vs-all network for universal domain adaptation},
  author={Saito, Kuniaki and Saenko, Kate},
  booktitle={Proceedings of the ieee/cvf international conference on computer vision},
  pages={9000--9009},
  year={2021}
}

@article{liang2021umad,
  title={Umad: Universal model adaptation under domain and category shift},
  author={Liang, Jian and Hu, Dapeng and Feng, Jiashi and He, Ran},
  journal={arXiv preprint arXiv:2112.08553},
  year={2021}
}

@inproceedings{wan2024unveiling,
  title={Unveiling the Unknown: Unleashing the Power of Unknown to Known in Open-Set Source-Free Domain Adaptation},
  author={Wan, Fuli and Zhao, Han and Yang, Xu and Deng, Cheng},
  booktitle={Proceedings of the IEEE/CVF Conference on Computer Vision and Pattern Recognition},
  pages={24015--24024},
  year={2024}
}

@article{fang2024source,
  title={Source-free unsupervised domain adaptation: A survey},
  author={Fang, Yuqi and Yap, Pew-Thian and Lin, Weili and Zhu, Hongtu and Liu, Mingxia},
  journal={Neural Networks},
  pages={106230},
  year={2024},
  publisher={Elsevier}
}

@inproceedings{saenko2010adapting,
  title={Adapting visual category models to new domains},
  author={Saenko, Kate and Kulis, Brian and Fritz, Mario and Darrell, Trevor},
  booktitle={Computer Vision--ECCV 2010: 11th European Conference on Computer Vision, Heraklion, Crete, Greece, September 5-11, 2010, Proceedings, Part IV 11},
  pages={213--226},
  year={2010},
  organization={Springer}
}

@inproceedings{venkateswara2017deep,
  title={Deep hashing network for unsupervised domain adaptation},
  author={Venkateswara, Hemanth and Eusebio, Jose and Chakraborty, Shayok and Panchanathan, Sethuraman},
  booktitle={Proceedings of the IEEE conference on computer vision and pattern recognition},
  pages={5018--5027},
  year={2017}
}

@article{peng2017visda,
  title={Visda: The visual domain adaptation challenge},
  author={Peng, Xingchao and Usman, Ben and Kaushik, Neela and Hoffman, Judy and Wang, Dequan and Saenko, Kate},
  journal={arXiv preprint arXiv:1710.06924},
  year={2017}
}

@inproceedings{peng2019moment,
  title={Moment matching for multi-source domain adaptation},
  author={Peng, Xingchao and Bai, Qinxun and Xia, Xide and Huang, Zijun and Saenko, Kate and Wang, Bo},
  booktitle={Proceedings of the IEEE/CVF international conference on computer vision},
  pages={1406--1415},
  year={2019}
}

@inproceedings{fu2020learning,
  title={Learning to detect open classes for universal domain adaptation},
  author={Fu, Bo and Cao, Zhangjie and Long, Mingsheng and Wang, Jianmin},
  booktitle={Computer Vision--ECCV 2020: 16th European Conference, Glasgow, UK, August 23--28, 2020, Proceedings, Part XV 16},
  pages={567--583},
  year={2020},
  organization={Springer}
}

@article{saito2020universal,
  title={Universal domain adaptation through self supervision},
  author={Saito, Kuniaki and Kim, Donghyun and Sclaroff, Stan and Saenko, Kate},
  journal={Advances in neural information processing systems},
  volume={33},
  pages={16282--16292},
  year={2020}
}

@inproceedings{li2021domain,
  title={Domain consensus clustering for universal domain adaptation},
  author={Li, Guangrui and Kang, Guoliang and Zhu, Yi and Wei, Yunchao and Yang, Yi},
  booktitle={Proceedings of the IEEE/CVF conference on computer vision and pattern recognition},
  pages={9757--9766},
  year={2021}
}

@inproceedings{chen2022geometric,
  title={Geometric anchor correspondence mining with uncertainty modeling for universal domain adaptation},
  author={Chen, Liang and Lou, Yihang and He, Jianzhong and Bai, Tao and Deng, Minghua},
  booktitle={Proceedings of the IEEE/CVF Conference on Computer Vision and Pattern Recognition},
  pages={16134--16143},
  year={2022}
}

@article{chang2022unified,
  title={Unified optimal transport framework for universal domain adaptation},
  author={Chang, Wanxing and Shi, Ye and Tuan, Hoang and Wang, Jingya},
  journal={Advances in Neural Information Processing Systems},
  volume={35},
  pages={29512--29524},
  year={2022}
}

@inproceedings{liang2020we,
  title={Do we really need to access the source data? source hypothesis transfer for unsupervised domain adaptation},
  author={Liang, Jian and Hu, Dapeng and Feng, Jiashi},
  booktitle={International conference on machine learning},
  pages={6028--6039},
  year={2020},
  organization={PMLR}
}

@inproceedings{long2018conditional,
  title={Conditional adversarial domain adaptation},
  author={Long, Mingsheng and Cao, Zhangjie and Wang, Jianmin and Jordan, Michael I},
  booktitle={Proceedings of the 32nd International Conference on Neural Information Processing Systems},
  pages={1647--1657},
  year={2018}
}

@inproceedings{zhang2019bridging,
  title={Bridging theory and algorithm for domain adaptation},
  author={Zhang, Yuchen and Liu, Tianle and Long, Mingsheng and Jordan, Michael},
  booktitle={International conference on machine learning},
  pages={7404--7413},
  year={2019},
  organization={PMLR}
}

@inproceedings{you2019universal,
  title={Universal domain adaptation},
  author={You, Kaichao and Long, Mingsheng and Cao, Zhangjie and Wang, Jianmin and Jordan, Michael I},
  booktitle={Proceedings of the IEEE/CVF conference on computer vision and pattern recognition},
  pages={2720--2729},
  year={2019}
}

@article{van2008visualizing,
  title={Visualizing data using t-SNE.},
  author={Van der Maaten, Laurens and Hinton, Geoffrey},
  journal={Journal of machine learning research},
  volume={9},
  number={11},
  year={2008}
}

@inproceedings{selvaraju2017grad,
  title={Grad-cam: Visual explanations from deep networks via gradient-based localization},
  author={Selvaraju, Ramprasaath R and Cogswell, Michael and Das, Abhishek and Vedantam, Ramakrishna and Parikh, Devi and Batra, Dhruv},
  booktitle={Proceedings of the IEEE international conference on computer vision},
  pages={618--626},
  year={2017}
}

\end{document}